\documentclass[runningheads]{llncs}

\usepackage{eccv}

\usepackage{eccvabbrv}
\usepackage{atbegshi}
\usepackage[dvipsnames]{xcolor}

\usepackage{graphicx}
\usepackage{booktabs}
\usepackage{microtype}
\usepackage{listings}
\usepackage[accsupp]{axessibility}
\usepackage[inline]{enumitem}
\usepackage{tcolorbox}
\usepackage{subcaption}
\usepackage{siunitx}
\usepackage{enumitem}
\usepackage{colortbl} 
\usepackage{wrapfig}
\usepackage{multirow} %
\usepackage{tabularx}
\tcbuselibrary{listings, breakable}

\newcommand{\suppmat}{\textit{Appendix\xspace}}
\newcommand{\datasetName}{ABO-Edit\xspace}

\def\xS{x_{\text{src}}}
\def\xT{x_{\text{trg}}}
\def\pStoT{p_{\text{src} \rightarrow \text{trg}}}
\def\triplet{$\langle\xS, \pStoT, \xT\rangle$\xspace}

\def\xzero{\mathbf{x}_{0}}
\newcommand{\hatxzero}{\hat{\mathbf{x}}_{0}}
\newcommand{\hatxzerot}{\hat{\mathbf{x}}_{0,t}}

\def\xone{\mathbf{x}_{1}}
\def\xt{\mathbf{x}_{t}}
\def\latent{\mathbf{z}}
\newcommand{\vPred}{\mathbf{v}_{\theta} (\xt, t)}

\newcommand{\hNoisyL}{\mathbf{h}_{L}^{\text{noisy}}}
\newcommand{\hCondL}{\mathbf{h}_{L}^{\text{cond}}}

\newcommand{\methodName}{FlowMirror\xspace}

\newcommand{\fid}{\texttt{FID}\xspace}
\newcommand{\ssim}{\texttt{SSIM}\xspace}
\newcommand{\lpips}{\texttt{LPIPS}\xspace}
\newcommand{\psnr}{\texttt{PSNR}\xspace}
\newcommand{\clip}{\texttt{CLIP}\xspace}
\newcommand{\dino}{\texttt{DINOv2}\xspace}

\newcommand{\mstd}[2]{#1{\tiny\textcolor{gray}{\,$\pm$#2}}}

\newcolumntype{G}{>{\columncolor[gray]{0.85}}S}

\usepackage[accsupp]{axessibility}  %

\usepackage[pagebackref,breaklinks,colorlinks,citecolor=eccvblue]{hyperref}
\usepackage{orcidlink}
\begin{document}
\title{DiTailed: Ensuring Visual Object Consistency in Text-Image-to-Image Flow Matching Models} 
\titlerunning{DiTailed: Ensuring Visual Object Consistency}
\author{Francesco Taioli\inst{1}\orcidlink{0000-0003-4133-4472}\and
Daniel Coelho\inst{1}\orcidlink{0000-0002-5743-7663}\and
Iaroslav Melekhov\inst{1}\orcidlink{0000-0003-3819-5280}\and \\
Roberto Alcover-Couso\inst{1}\orcidlink{0000-0001-9609-4416}\and 
Jose Miguel Grande Saiz\inst{1}\orcidlink{0009-0000-5822-1154} \and \\
Virginia Fernandez Arguedas\inst{1} \and
Artur Bekasov \inst{2,}\thanks{Work done while at Amazon.}}
\authorrunning{F.~Taioli et al.}
{\renewcommand{\and}{\hspace{2em}}%
\institute{Amazon, \email{\{ftaioli,dfcoelho,imlkhv,ralcover,jgrande,virfer\}@amazon.com} \and
Faculty, \email{artur.bekasov@faculty.ai}}}
\maketitle
\begin{figure}[h]
    \begin{center}
    \includegraphics[width=0.98\textwidth]{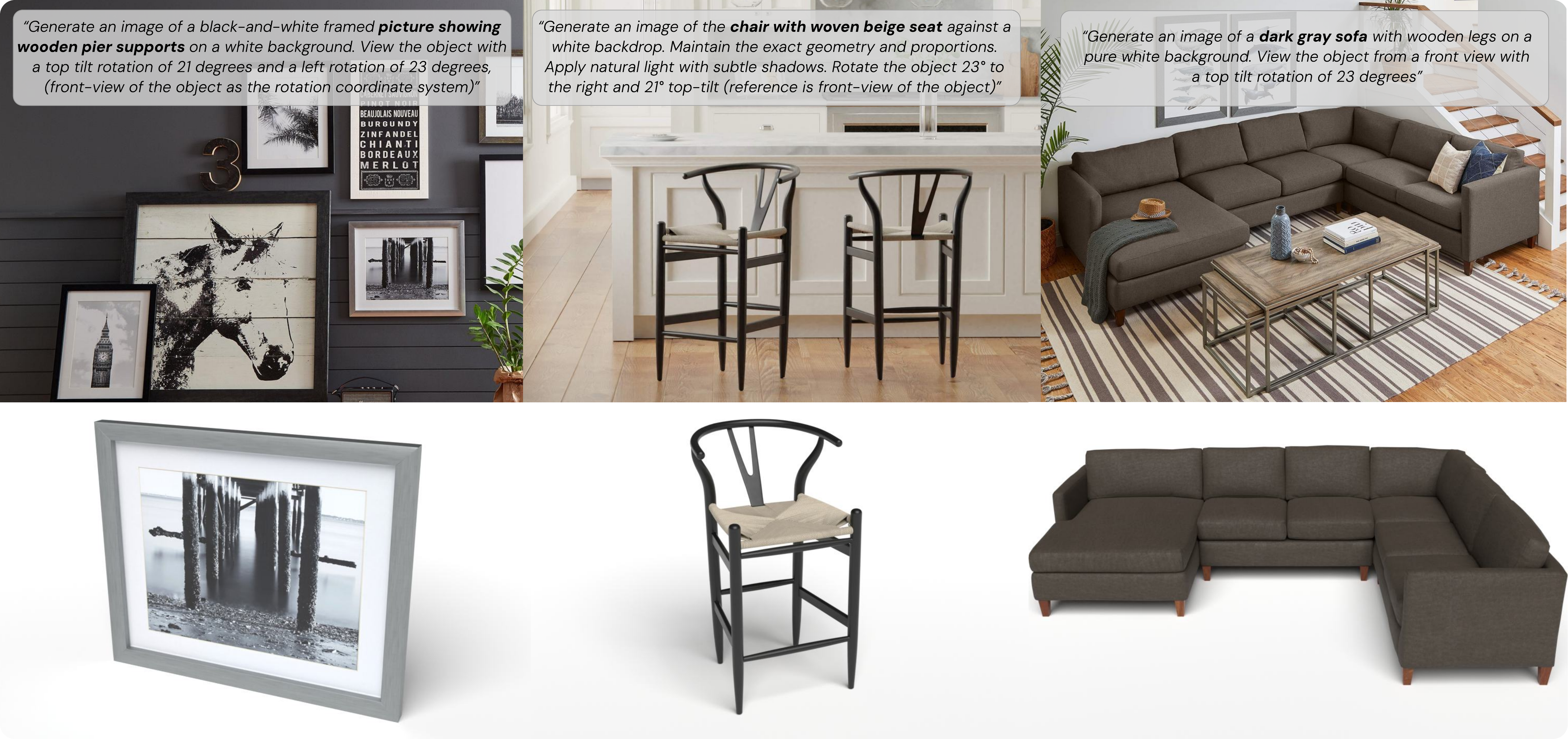}
    \caption{\textit{\textbf{\datasetName}} benchmarks \textit{visual consistency}: the degree to which editing models preserve an object's visual attributes, \eg, texture, color, geometry, and fine-grained details.
    Spanning $12\text{K+}$ samples over diverse objects and edits, it systematically probes whether these attributes survive changes in viewpoint, context, and appearance.}
    \label{fig:teaser}
    \end{center}
\end{figure}

\vspace{-0.6cm}
\begin{abstract}
Despite remarkable progress in text-guided image editing, generative models frequently fail to preserve visual object consistency, defined as the preservation of a subject's key attributes throughout the editing process.
We address this limitation through three contributions.
First, we introduce \datasetName, a dataset specifically designed to study object consistency, comprising over $12\text{,}000$ triplets of source images, editing prompts, and high-quality target images rendered from artist-designed 3D assets, with multi-view coverage and human-verified quality control.
Second, we uncover an overlooked property of image-editing rectified flow models: the conditioning embedding space, not directly supervised during training, encodes a prediction of the final generated image even at high noise levels. 
Third, exploiting this finding, we propose \methodName, a parameter-free auxiliary loss that supervises this conditioning embedding space.
Without architectural changes, our method improves generation quality across several metrics over baselines. Page: \href{https://francescotaioli.github.io/DiTailed}{francescotaioli.github.io/DiTailed}
\end{abstract}

\section{Introduction}
\label{sec:intro}
The advent of large-scale generative models~\cite{nano_banana,nano_banana_pro,qwen_image_edit, flux2, flux_1,sdxl} has revolutionized the field of image synthesis, enabling the creation of high-fidelity images from diverse input modalities.
These capabilities could potentially show value across various domains, including creative design~\cite{Wang_2025_CVPR}, digital media production, and content creation~\cite{Lin_2023_CVPR}.

Despite these significant advances, several challenges remain unresolved.
\begin{figure*}[t]
  \centering
  \includegraphics[width=\linewidth]{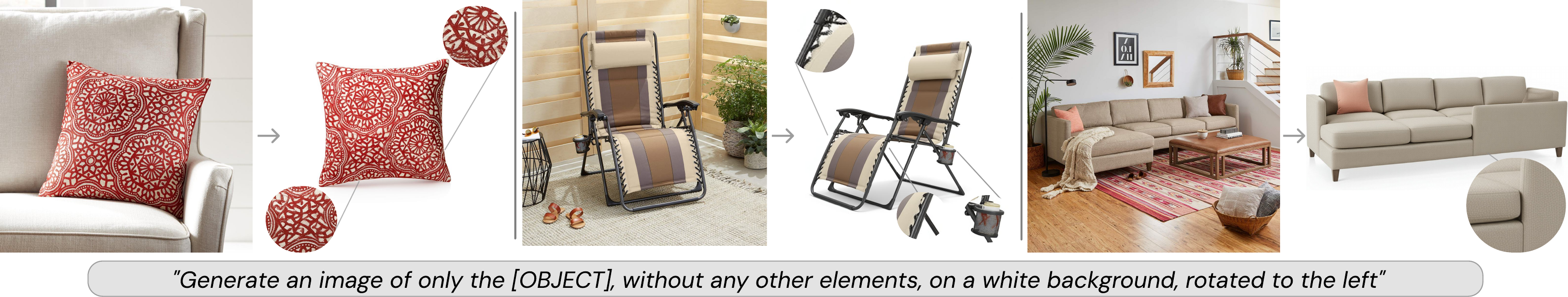}
  \caption{\textit{Visual  object inconsistencies} observed in current image-editing models: (\textit{left}) pillow patterns are not preserved, (\textit{center}) cup holder and rivets are missing or poorly reconstructed, and (\textit{right}) sofa geometry is entirely altered with inconsistent texture.}
  \label{fig:intro_failure_cases}
\end{figure*}
Even with seemingly straightforward prompts specifying background removal or basic geometric transformations (\eg, rotation), state-of-the-art generative models frequently fail to maintain \textit{visual object consistency}~\cite{Cao_2023_ICCV,hertz2023prompttoprompt,1_to_3d,Yu_2023_ICCV}, defined as the preservation of an object's intrinsic visual attributes, including geometry, texture, color, and fine-grained details, throughout the image generation process, except where explicitly requested by the conditioning prompt.

We illustrate such failure cases in Fig.~\ref{fig:intro_failure_cases}, where current large-scale open-weights image generation foundation models can struggle to keep the object consistent.
As shown, when prompted to rotate an object and remove the background, the model fails to preserve critical visual details: the pattern on the pillow is altered, the support rings and cup-holders are missing or poorly generated, and the sofa's shape is substantially distorted with a texture that misrepresents the original product.
This lack of fine-grained element preservation may limit the applicability of current generative models for downstream applications that could demand pixel-level visual fidelity, such as product visualization, 3D digital content creation~\cite{1_to_3d}, novel view synthesis~\cite{Yu_2023_ICCV}, and product standardization~\cite{automati_gen}.

To address these limitations, we make three contributions: a dedicated benchmark for measuring visual object consistency, novel analytical insights into the internal behavior of rectified flow editing models, and a parameter-free auxiliary loss that leverages these insights to improve generation quality.

First, we introduce \textit{\datasetName}, a curated benchmark for visual object consistency built on the Amazon Berkeley Objects (ABO) dataset~\cite{abo_dataset}\footnote{Amazon Berkeley Objects (ABO) dataset is publicly available under CC BY 4.0 License at \url{https://amazon-berkeley-objects.s3.amazonaws.com/}.}.
As shown in Fig.~\ref{fig:teaser}, \datasetName contains triplets consisting of:
\begin{enumerate*}[label=\textit{(\roman*)}]
    \item a high-definition source image;
    \item a textual prompt specifying the intended transformation, and 
    \item a high-definition target image representing the desired output.
\end{enumerate*}
Source images depict products in ``lifestyle'' settings, \ie, real-world scenes with complex backgrounds, occlusions, and contextual clutter, posing a particularly challenging scenario for generative models that must disentangle object-specific features from high visual noise.
\datasetName covers a diverse set of product types, including furniture (\eg, sofas, chairs), wall art, and decorative items.
To ensure ground-truth object consistency, target images are rendered in high resolution (1024$\times$1024) from the corresponding artist-designed $3$D models  using Blender~\cite{blender}, guaranteeing accurate preservation of object geometry, texture, and fine-grained details. 
We render target images on a white background, with realistic shadows.
Furthermore, our benchmark includes multi-view rendering of each object (\ie, \textit{left}, \textit{front} and \textit{right} views) at different rotation angles, which allows evaluating how well generative models maintain object consistency under object orientation transformations.

Second, we provide novel insights into the behavior of state-of-the-art image-editing models~\cite{kanidsky5,qwen_image_edit, flux_1, flux2}\footnote{All evaluated models are open-weight and publicly available.} during the editing process.
In particular, we identify that a specific model family already encodes a latent prediction of the final output in the conditioning embedding space of the input image, even at high-noise levels.
Remarkably, if supervised, this latent prediction exhibits higher similarity to the target than the velocity estimate at early denoising steps, providing insight into the effective number of inference steps required for convergence.

Building on these findings, we propose a parameter-free auxiliary loss that exploits this intrinsic predictive behavior to enhance object consistency during generation, showing improvements in generation quality across multiple metrics and architectures.
In summary, this paper makes the following contributions:
\begin{enumerate}[label=\roman*., align=right]
  \item We introduce \textit{\datasetName}, a benchmark for training and evaluating object consistency in image editing, featuring $12k+$ high-quality human-reviewed source-prompt-target triplets with ground-truth 3D renderings and degree-level orientation annotations. The dataset will be released upon acceptance;
  \item We identify that rectified flow editing models encode target predictions in the conditioning embedding space even at high noise levels; and
  \item We propose \textit{\methodName}, a parameter-free auxiliary loss that supervises this embedding space to improve generation quality across metrics over baselines.
\end{enumerate}

\section{Related Works}
\label{sec:related_works}
This section reviews generative models; related datasets in Sec.~\ref{sup:dataset:related_work_dataset} (\suppmat).
\noindent\textbf{Diffusion Models.}
Diffusion models~\cite{diffusion, ddpm, diffusion_survey} have emerged as a powerful class of generative models, achieving state-of-the-art results in modeling high-dimensional data distributions.
While early efforts demonstrated strong performance, their widespread adoption was enabled by the computational efficiency gained through operating in a compressed latent space~\cite{sd1}. Subsequent advances in classifier-free guidance~\cite{cfg}, deterministic sampling~\cite{ddim}, and architectural design~\cite{diffusion_design} have further enhanced both sample quality and practical deployment.
These models have revolutionized generative modeling with applications spanning image synthesis~\cite{ddpm, sd1, ramesh2022hierarchicaltextconditionalimagegeneration, sdxl}, video generation~\cite{video_diff_model, ho2022imagenvideohighdefinition}, and 3D generation~\cite{poole2023dreamfusion}.

\noindent\textbf{Rectified Flow Models.}
Rectified flow models~\cite{lipman2023flow, flow_straight_and_fast} simplify the generative process by learning straight trajectories between noise and data distributions, enabling more efficient sampling with fewer evaluation steps through linear probability paths.
Concurrently, architectural advances have shifted from U-Net-based architectures to transformer-based designs.
The Diffusion Transformer (DiT)~\cite{dit} demonstrated that the classic transformer architecture can effectively replace the standard U-Net architecture, showing promising scaling properties.
Furthermore, the MultiModal Diffusion Transformer (MM-DiT)~\cite{mmdit} extended this approach to jointly model multiple modalities, \eg, with images and text.
Modern rectified flow models leverage powerful pretrained text encoders for conditioning. 
SD3~\cite{mmdit} and Flux~\cite{flux_1} employ a combination of CLIP~\cite{clip} and T5~\cite{t5} encoders to capture both semantic alignment and fine-grained textual understanding, while Qwen-Image~\cite{qwen_image_edit} obtains semantic embeddings from a VLM~\cite{qwen_2_5_vl}.
Building on these foundations, recent works have demonstrated the effectiveness of large-scale, rectified flow models for video and image synthesis~\cite{flux_1, flux2, qwen_image_edit,z_image,hiDream,in_context_edit, sdxl,wan2025wanopenadvancedlargescale,nano_banana,nano_banana_pro}, establishing new benchmarks for visual generation quality.

\section{Dataset}
\label{sec:dataset}
We introduce \datasetName, a curated dataset for training and evaluating generative models on \textit{Visual Object Consistency}.
\datasetName addresses the challenging task of transforming \textit{``Lifestyle''} images (depicting products in complex real-world usage scenarios) into \textit{studio-quality} representations: the same product isolated on a white background with realistic shadow, rotated and tilted to a precisely specified angle (see~Fig.~\ref{fig:teaser}).
This task addresses a general challenge in e-commerce platforms~\cite{automati_gen,thousand_word}, where standardized and high-fidelity product imagery is commonly desired. 
Prior research in e-commerce has shown that \textit{inconsistent} product imagery may affect user experience and search effectiveness~\cite{Brylla01102020,Maier02012019}.
Specifically, each sample comprises a triplet \triplet,
where $\xS$ denotes a source lifestyle image, $\pStoT$ represents a detailed editing prompt including fine-grained rotation angles, and $\xT$ is the corresponding ground-truth target image rendered from a 3D asset.
Beyond the editing task evaluated here, \datasetName supports the reverse task (\textit{studio-to-lifestyle} generation), single-image 3D reconstruction, and text-to-3D generation.

\subsection{Dataset Construction}
Our dataset is built upon the Amazon Berkeley Objects (ABO) Dataset~\cite{abo_dataset}, a large-scale, publicly available collection of high-quality images depicting household objects.
ABO provides, for more than $7\text{,}900$ products, high-quality, artist-created 3D models with 4K texture maps and physically-based materials. 
In the following, we describe how we construct \datasetName.

\subsection{Filtering procedure}
We implement a two-stage automatic filtering pipeline, described below.

\noindent\textbf{Product-level filtering.}
To ensure high-quality $\langle\xS, \xT\rangle$\xspace pairs, we implement an automatic filtering pipeline, described as follows.
We first exclude:~
\begin{enumerate*}[label=\textit{(\roman*)}]
    \item product types with minimal transformation requirements (\eg, phone cases);
    \item product types not associated with 3D models (ABO~\cite{abo_dataset} provides 3D models for more than $7\text{,}900$ products); and
    \item product types where rendering is challenging due to material properties (\eg, mirrors, transparent objects).
\end{enumerate*}

\noindent\textbf{Image-level filtering.}
We then discard source images $\xS$ lacking ``\textit{lifestyle}'' context. 
To automate this process, we curate a small annotated dataset and train an MLP classifier on CLIP~\cite{clip} embeddings to distinguish lifestyle images from studio-shot or plain product images.
The classifier achieves $90\%$ precision and $82\%$ recall.
Images passing this filter are retained as source images $\xS$.

\subsection{Target Image Generation}
To create corresponding target images $\xT$, we leverage the 3D models associated with each product sample. 
Specifically, using the Blender Python API\footnote{Blender is available under the GNU GPL License at \url{https://www.blender.org/}.}, we render high-quality RGB images of the products at $1024\times1024$ resolution.

Notably, for each source image $\xS$, we generate three target images $\xT$ corresponding to \textit{front}, \textit{left}, and \textit{right} views of the object. 
For each view, we define a valid range of camera angles and randomly sample both an azimuth angle and an elevation angle relative to the object's front-facing coordinate system. 
The camera is then positioned accordingly, ensuring the object remains within the field of view. We show samples in the \suppmat, Fig.~\ref{fig:sup:rotation_information}.

\subsection{Prompt Generation}
We now describe the generation of the prompt $\pStoT$, which specifies the desired transformation from source to target image.
We begin with a static template, shared across all samples, which we populate with sample-specific information. 
The full template is provided in the \suppmat~(Sec.~\ref{sup:dataset:prompt:static_prompt_dataset}).

Specifically, we populate the template with two sample-specific fields:
\begin{enumerate*}[label=\textit{(\roman*)}]
\item \textit{Object category and description}: we employ a VLM to generate a detailed description of the object and extract its category from $\xT$ (the product is rendered on a white background). 
Including a fine-grained description enables the editing model to disambiguate the target object when multiple objects of the same category are present in the source image.
A manual inspection of $100$ randomly sampled triplets revealed only $4$ minor counting inaccuracies (\eg, number of sofa seats or drawers); and

\item \textit{Rotation data}: we encode the ground-truth camera rotation from Blender as a formatted string.
We promote output diversity by selecting different system prompts and instruction wording during prompt construction.
\end{enumerate*}
\subsection{Human Quality Control}
While the rendering process itself is deterministic, the resulting triplets may still contain quality issues that are difficult to detect automatically: 
\textit{(i)}~objects in the source image may be partially cropped or occluded, making attribute-preserving editing ill-defined;
\textit{(ii)}~programmatic light placement in Blender can produce over- or under-exposed renders depending on material properties;
and~\textit{(iii)}~the product depicted in the source image may not correspond to the associated 3D model.
We therefore perform manual quality control via a crowdsourcing platform to validate the final triplets.
Further details are provided in the \suppmat~(Sec.~\ref{sup:dataset:human_quality_control}).

\begin{figure*}[t]
  \centering
  \includegraphics[width=\linewidth]{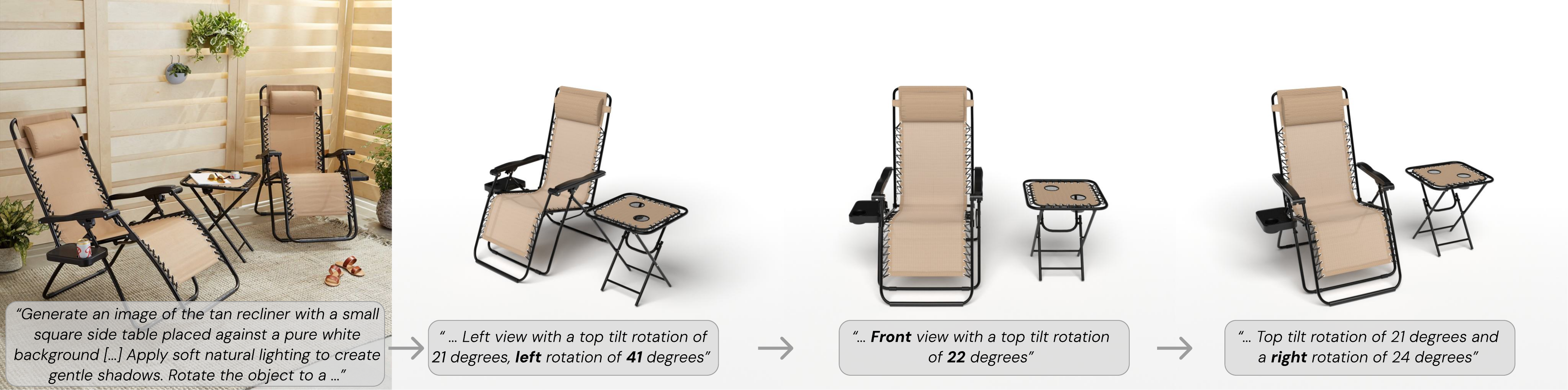}
  \caption{A single 3D object rendered from three canonical viewpoints: \textit{left},
    \textit{front}, and \textit{right}. Each view includes the corresponding 
    text prompt specifying the target orientation, along with precise elevation and 
    azimuth annotations.}
  \label{fig:rotation_information_single_example}
\end{figure*}
We summarize the dataset statistics in Table~\ref{tab:statistics}.
\subsection{Statistics}
\datasetName comprises $12{,}319$ training and $400$ validation samples spanning $20$ product families.
Chairs, sofas, and tables represent the most prevalent categories, with $2{,}800$, $1{,}821$, and $1{,}168$ samples, respectively (see \suppmat~Fig.~\ref{fig:sup:product_distribution}).
\begin{wraptable}[12]{r}{0.58\textwidth}
    \centering
    \caption{\textbf{\datasetName statistics.} We report the number of samples and the distribution of camera pose parameters (mean $\pm$ std) used during image rendering. Elevation is sampled uniformly within a fixed range, while azimuth varies by view direction.}
\begin{tabular}{lcc}
\toprule
\textbf{Statistic} & \textbf{Train} & \textbf{Validation} \\
\midrule
Number of samples
    & 12{,}319 & 400 \\
Elevation angle
    & {\mstd{\SI{22.0}{\degree}}{0.8}} & {\mstd{\SI{22.0}{\degree}}{0.8}} \\
Azimuth (left view)
    & {\mstd{\SI{32.3}{\degree}}{7.2}} & {\mstd{\SI{32.5}{\degree}}{7.5}} \\
Azimuth (right view)
    & {\mstd{\SI{15.4}{\degree}}{5.8}} & {\mstd{\SI{15.2}{\degree}}{5.5}} \\
\bottomrule
\label{tab:statistics}
\end{tabular}

\end{wraptable}
\noindent\textbf{Camera Pose Distribution.}
For each source image, we render three target views (front, left, and right) by positioning the camera at randomly sampled poses relative to the object's front-facing coordinate system. 
Specifically, the camera \emph{elevation} (vertical angle above the horizontal plane) is sampled around \SI{22}{\degree}, providing a slight top-down perspective.
The camera \emph{azimuth} (horizontal rotation around the object) varies by view: front views use \SI{0}{\degree},
left views average \SI{32.2}{\degree}, while right views average \SI{15.4}{\degree} in the opposite direction (shown as absolute value in Table~\ref{tab:statistics}).
We provide a visual example in Fig.~\ref{fig:rotation_information_single_example}.
This asymmetry in azimuth ranges encourages diverse viewpoint coverage while maintaining natural viewing angles.
Further details on product and rotation angle distributions are provided in the \suppmat~(Sec.~\ref{sup:dataset:product_types_distribution} and Fig.~\ref{fig:sup:rotation_distribution}).

\section{Method}
\label{sec:method}
\subsection{Preliminaries}
\noindent\textbf{Flow Models}
aim to transform a simple initial distribution $p_{1}$ into a complex data distribution $p_{0}$\footnote{We use the inverted time convention, where $t = 0$ corresponds to a data sample.}.
Formally, let $\xzero \sim p_0$ denote a \textit{data sample}, and $\xone \sim p_1$ denote a sample from the initial distribution, \ie, the standard multivariate normal distribution $\mathcal{N}(0, \mathbf{I})$.
A flow model defines a time-dependent probability path $p_t$ for $t \in [0,1]$ that interpolates between $p_1$ and $p_0$, governed by an ordinary differential equation (ODE):
\begin{equation}
\frac{\partial \xt }{\partial t} = u_{t}^{\theta}(\xt),
\end{equation}
where $u^\theta_t(\xt)$ is a time-dependent velocity field parameterized by a neural network.
Sampling is performed by integrating this ODE from $t=1$ to $t=0$.

\noindent\textbf{Rectified Flow}~\cite{lipman2023flow, flow_straight_and_fast} considers a linear probability path induced by interpolating between samples $\xone$ and $\xzero$, yielding a constant velocity field:

\noindent%
\begin{minipage}{0.48\linewidth}
\begin{equation}
    \xt = (1-t)\,\xzero + t\,\xone,
    \label{eq:linear_interpolation}
\end{equation}
\end{minipage}%
\hfill
\begin{minipage}{0.48\linewidth}
\begin{equation}
    \mathbf{v} = \tfrac{\partial \xt}{\partial t} = \xone - \xzero.
    \label{eg:constant_velocity}
\end{equation}
\end{minipage}

The neural network predicts velocity $\vPred$, approximating the velocity field by minimizing the Conditional Flow Matching objective (CFM):
\begin{equation}
\mathcal{L}_{\mathrm{CFM}}(\theta)
=
\mathbb{E}_{\xzero \sim p_0,\, \xone \sim p_1,\, t \sim \mathcal{U}[0,1]}
\left[
\left\lVert
\vPred - (\xone -\xzero)
\right\rVert^2
\right].
\label{eg:cfm}
\end{equation}
\noindent\textbf{MultiModal Diffusion Transformer (MM-DiT)}~\cite{mmdit} extends the DiT architecture~\cite{dit} by jointly modeling text and images in a shared latent space.
Let $\mathbf{x} \in \mathbb{R}^{H \times W \times 3}$ denote the target RGB image.
A pretrained VAE encoder $\mathcal{E}$ maps the image $\mathbf{x}$ to a latent representation $\latent = \mathcal{E}(\mathbf{x}) \in \mathbb{R}^{h \times w \times c}$, where $h = H/r$, $w = W/r$,  $r$ denotes the spatial compression rate (typically $8$), and $c$ is the latent channel dimension.
The latent $\latent$ is then packed by merging $2 \times 2$ spatial patches into 
the channel dimension, yielding image tokens $\latent_{\text{img}} \in 
\mathbb{R}^{N_{\text{img}} \times d}$, where $N_{\text{img}}$ is the number of image tokens and $d$ is the hidden dimension.
Similarly, a text prompt is encoded using pretrained text encoders (\eg,~\cite{clip, t5}) to obtain text tokens $\latent_{\text{txt}} \in \mathbb{R}^{N_{\text{txt}} \times d}$, where $N_{\text{txt}}$ is the sequence length.
The noisy latent $\latent_{\text{noisy}} \in \mathbb{R}^{N_{\text{img}} \times d}$ is constructed from $\latent_{\text{img}}$ following Eq.~\ref{eq:linear_interpolation}.

For image-editing tasks, MM-DiT additionally conditions on an input RGB image 
$\mathbf{x}_{\text{cond}} \in \mathbb{R}^{H \times W \times 3}$.
This conditioning image is processed through the same VAE encoder and packing 
operation to obtain latent $\latent_{\text{cond}} \in \mathbb{R}^{N_{\text{cond}} \times d}$.
The noisy and conditioning tokens are then concatenated along the sequence dimension to form the transformer input, and then processed through $L$ transformer blocks:
\begin{align}
    \mathbf{h}_0 &= [\latent_{\text{noisy}}; \latent_{\text{cond}}] 
        \in \mathbb{R}^{(N_{\text{img}} + N_{\text{cond}}) \times d}, 
        \label{eq:concat} \\
    \mathbf{h}_\ell &= \text{TransformerBlock}_\ell(
        \mathbf{h}_{\ell-1}, \latent_{\text{text}}, t), 
        \quad \ell = 1, \dots, L,
        \label{eq:transformer}
\end{align}
where $t$ is the diffusion timestep.
The last hidden state $\mathbf{h}_L \in \mathbb{R}^{(N_{\text{img}} + N_{\text{cond}}) \times d}$ is split along the sequence dimension to recover the noisy and conditioning components:
\begin{equation}
\mathbf{h}_L = [\hNoisyL; \hCondL], \quad \text{where } \hNoisyL \in \mathbb{R}^{N_{\text{img}} \times d}, \; \hCondL \in \mathbb{R}^{N_{\text{cond}} \times d}.
\end{equation}
Finally, the velocity prediction $\mathbf{v}$ is computed solely from the noisy tokens via a learned projection:
\begin{equation}
\mathbf{v} = \text{Proj}(\hNoisyL).
\end{equation}

\subsection{Observations}
The predicted velocity $\mathbf{v}$ is computed solely from $\hNoisyL$, while the conditioning output $\hCondL$ is discarded at inference.
Yet $\hCondL$ remains coupled to the generation process: it participates in attention computations across all transformer layers and receives gradients during training.
This raises a question: \emph{what information does $\hCondL$ encode, and how does it evolve during denoising?}
We investigate this through two complementary analyses.

\begin{figure*}[t]
  \centering
  \includegraphics[width=\linewidth]{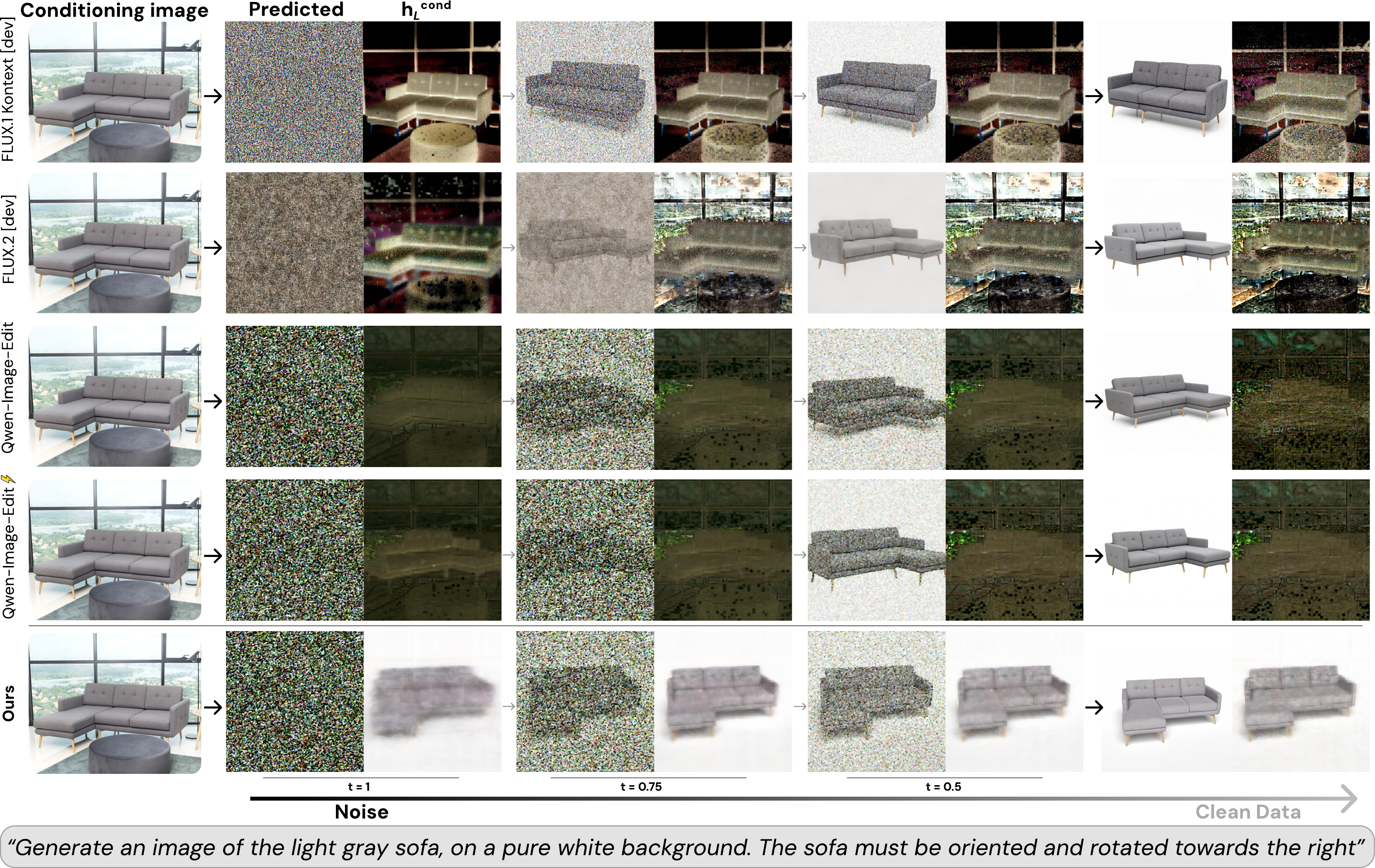}
  \caption[Visualization of internal representations across different models]{
  \textbf{Internal representations.}
  We visualize both the predicted image and the decoded conditioning latent $\hCondL$ at different timesteps during denoising.
  Model evaluated \textit{without} fine-tuning:~
  \begin{enumerate*}[label=\textit{(\roman*)}]
  \item \texttt{FLUX.1 Kontext [dev]}~\cite{flux_1};
  \item \texttt{FLUX.2 [dev]}~\cite{flux2};
  \item \texttt{Qwen-Image-Edit-2509-Lightning}~\cite{qwen_image_lightning} and
  \item \texttt{Qwen-Image-Edit-2509}~\cite{qwen_image_edit}.
  \end{enumerate*}
  We also visualize our method's internal representation after fine-tuning.
  FLUX models maintain representations of the conditioning image throughout 
  denoising, whereas Qwen models \textit{encode the final prediction} in $\hCondL$ even at 
  high noise levels ($t \approx 1$). 
  Note that our method encodes the prediction in the
  embedding space \textit{while} preserving the original geometry.}
  \label{fig:image_internal_representation}
\end{figure*}

\noindent\textbf{Observation 1: Internal Representation.}
Fig.~\ref{fig:image_internal_representation} reveals striking differences in how state-of-the-art image editing models utilize the conditioning latent $\hCondL$ throughout the denoising process.
We decode $\hCondL$ using the VAE decoder $\mathcal{D}$ at various timesteps $t \in [0,1]$, and observe two distinct behavioral patterns. 

Models in the FLUX family~\cite{flux_1, flux2} maintain representations of the conditioning image $\mathbf{x}_{\text{cond}}$ throughout denoising. 
In contrast, Qwen-based models~\cite{qwen_image_edit, qwen_image_lightning} encode 
a prediction of the \emph{final generated image} in $\hCondL$, even at high noise 
levels ($t \approx 1$). 
This behavior is particularly pronounced in Qwen-Image-Edit-2509-Lightning~\cite{qwen_image_lightning}, a distilled, $8$-step variant of Qwen-Image-Edit-2509~\cite{qwen_image_edit}.

\noindent\textbf{Observation 2: Velocity Prediction.}
From Eq.~\ref{eg:constant_velocity}, we observe that rectified flow models predict a time-dependent velocity field corresponding to the displacement between data and noise along the linear interpolation path.
At any time step $t$, we can recover an \textit{estimate} of the clean image $\hatxzerot$ from the current state $\xt$ via
\begin{equation}
\hatxzerot = \xt - t\vPred.
\label{eq:x0_pred}
\end{equation}
We provide the derivation in Sec.~\ref{supmat:x0_derivation} (\suppmat).

\begin{figure*}[t]
  \centering
  \includegraphics[width=1\linewidth]{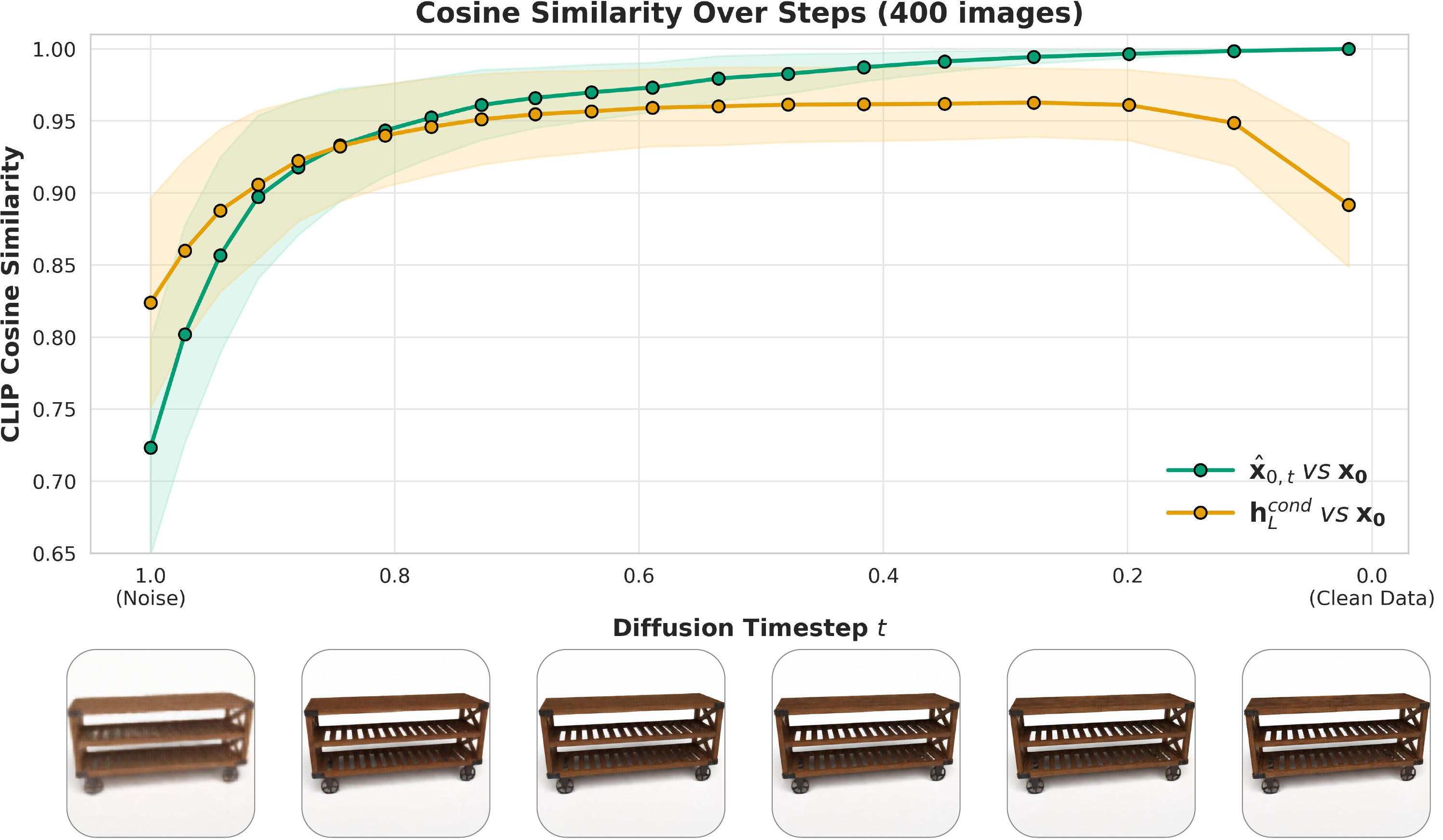}
    \caption{\textbf{Temporal evolution of predictions during denoising.}
    Mean cosine similarity using CLIP embeddings on 
  \datasetName{} validation set ($n=400$).
    \textcolor{OliveGreen}{\textbf{Green}}:~similarity between \textit{intermediate} prediction $\hatxzerot$ at timestep $t$ and \textit{final} output $\hatxzero$.
      \textcolor{Orange}{\textbf{Orange}}: similarity between decoded conditioning 
      embedding $\hCondL$ at timestep $t$ and final output $\hatxzero$ (boxes show a decoded sample at varying $t$).
    High similarity at early timesteps ($t \approx 1$) indicates that intermediate predictions already approximate the final generation.}
  \label{fig:similarity_vs_step}
\end{figure*}

Following Eq.~\ref{eq:x0_pred}, in Fig.~\ref{fig:similarity_vs_step} (green curve) we plot the cosine similarity between the predicted clean image $\hatxzerot $ at timesteps $t$ and the final generated image $\hat{\mathbf{x}}_0$.
Embeddings are computed using the vision backbone of \texttt{CLIP-ViT-L/14}~\cite{clip}.
Even at high noise levels ($t = 1$), the similarity remains relatively high, indicating that the predicted velocity at early timesteps already closely  approximates the final velocity.
Note that this similarity naturally converges to $1$ as $t \to 0$, since $\hatxzerot$ approaches $\hatxzero$ by construction.
A similar finding is reported in~\cite{li2025treft}.

\subsection{\methodName: When Conditioning Mirrors the Generation}
\begin{figure*}[h]
  \centering
  \includegraphics[width=\linewidth]{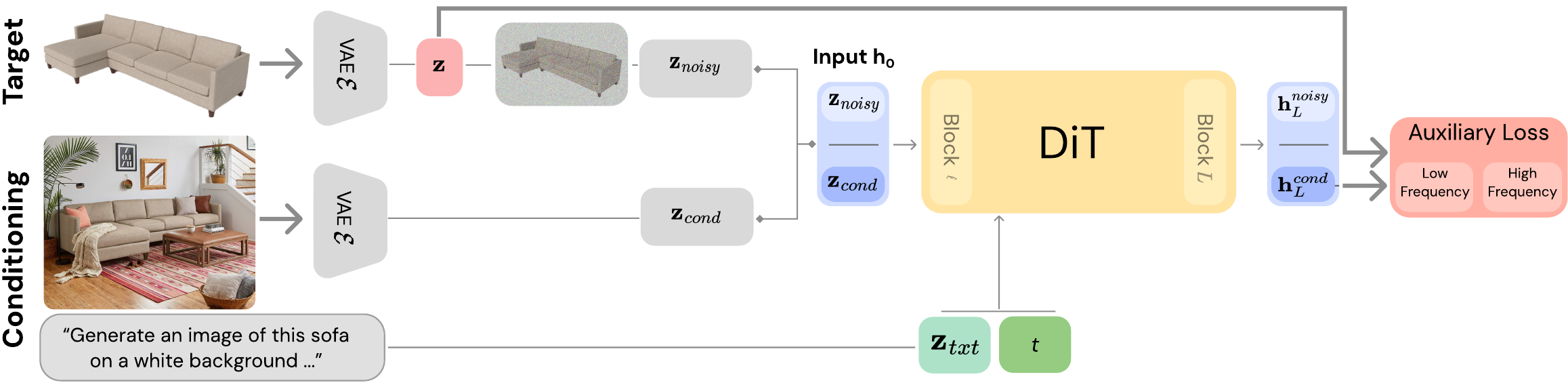}
  \caption{\textbf{Overview of the proposed auxiliary loss.}
      Given noisy input $\latent_{\text{noisy}}$ (\textit{target} image), conditioning image embedding $\latent_{\text{cond}}$, timestep $t$ and text embeddings $\latent_\text{txt}$, the transformer (DiT) predicts velocity $v$ while also producing the conditioning output $\hCondL$.
    We supervise $\hCondL$ against the target latent $\latent$ using a parameter-free, multi-scale decomposition that captures both fine details and global structure.}
  \label{fig:method_overview}
\end{figure*}
\noindent\textbf{Motivation.}
\textit{Observation}~\textbf{1} reveals that certain model families naturally encode the final output in 
$\hCondL$, while \textit{Observation}~\textbf{2} shows that output predictions stabilize early 
in denoising.
We hypothesize that explicitly supervising $\hCondL$ to match the target amplifies this beneficial behavior: it encourages the model to form accurate output representations early, while providing auxiliary gradient signal without additional parameters.

Based on these findings, we propose \textit{\methodName}, a regularization 
objective that supervises the conditioning output $\hCondL$, typically not enforced 
during training.
Crucially, our approach is \emph{parameter-free} and incurs negligible 
computational overhead.
Fig.~\ref{fig:method_overview} illustrates the overall framework.

\noindent\textbf{Overview.}
Our key idea is to supervise the transformer's conditioning output $\hCondL$
using the VAE latent representation of the \textit{target} image, $\latent$.
Operating in the VAE's compressed embedding space offers two advantages:
\begin{enumerate*}[label=\textit{(\roman*)}]
\item the lower-dimensional representation is computationally efficient, and
\item  all operations remain in latent space without invoking the decoder $\mathcal{D}$, avoiding substantial overhead.
\end{enumerate*}

To enable this supervision, we first reshape (``unpack'')  both representations into a shared spatial layout: $\latent \in \mathbb{R}^{c \times h \times w}$ and $\hCondL \in \mathbb{R}^{c \times h \times w}$.

\noindent\textbf{Multi-Scale Decomposition.}
We construct a Laplacian-pyramid–style decomposition over $S$ scales to capture both fine details and global structure.
Let $\mathbf{\hCondL}^{(s)}$ and $\mathbf{z}^{(s)}$ denote representations at scale $s$, where each level applies a fixed, depthwise Gaussian blur $\mathcal{G}(\cdot)$ independently per channel.

\noindent\textbf{High-Frequency (Detail Preservation)}.
To encourage detail preservation, we penalize the $\ell_1$ loss between the original and blurred tensors:
\begin{equation}
\mathcal{L}_{\mathrm{HF}}^{(s)} =
\left\|
\bigl(\mathbf{\hCondL}^{(s)} - \mathcal{G}(\mathbf{\hCondL}^{(s)})\bigr)
-
\bigl(\mathbf{z}^{(s)} - \mathcal{G}(\mathbf{z}^{(s)})\bigr)
\right\|_{1}
\end{equation}
\noindent\textbf{Low-Frequency (Structure)}.
Low-frequency components capture global structure.
After Gaussian blurring, we downsample, via average pooling $\mathrm{Down}(\cdot)$, and penalize discrepancy using an $\ell_2$ loss:
\begin{equation}
\mathcal{L}_{\mathrm{LF}}^{(s)} =
\left\|
\mathrm{Down}\!\left(\mathcal{G}(\mathbf{\hCondL}^{(s)})\right)
-
\mathrm{Down}\!\left(\mathcal{G}(\mathbf{z}^{(s)})\right)
\right\|_{2}^{2}
\end{equation}
\noindent\textbf{Total Loss.}
The full objective combines the standard CFM loss (Eq.~\ref{eg:cfm}) with our proposed auxiliary regularization:
\begin{equation}
\mathcal{L}_{\mathrm{total}} =
\mathcal{L}_{\mathrm{CFM}} + \lambda_{\mathrm{aux}} \, \mathcal{L}_{\mathrm{aux}}, \qquad
\mathcal{L}_{\mathrm{aux}} = \tfrac{1}{S}\sum_{s=1}^{S}
\left(
\mathcal{L}_{\mathrm{LF}}^{(s)} +
\lambda_{\mathrm{HF}} \cdot \mathcal{L}_{\mathrm{HF}}^{(s)}
\right).
\end{equation}

\section{Experiment}
\noindent\textbf{Dataset.}
We evaluate on \datasetName, which provides challenging evaluation scenarios for visual object consistency, including:
\begin{enumerate*}[label=\textit{(\roman*)}]
\item precise viewpoint control with degree-level angular accuracy;
\item $3$D spatial transformations such as rotation and tilting;
\item complex image editing requests combining product selection, background removal and pose request; and
\item object-prompt alignment across visually similar products.
\end{enumerate*}

\noindent\textbf{Implementation Details.}
We generate images at $1024 \times 1024$ resolution with each model's default configuration unless otherwise specified.
Due to computational constraints and a resource-constrained setting, we train with a budget of $10$k optimization steps on a single node equipped with $8$ NVIDIA H$100$ GPUs, using an effective batch size of $32$.
We fine-tune using DoRA~\cite{liu2024dora} with rank $r=64$, learning rate $\eta = 10^{-4}$ and scaling factor $\alpha=64$, maintaining a training resolution of $1024 \times 1024$.
Training completes in approximately $43$ hours ($336$ GPU-hours). 
For fine-tuned models, we evaluate three checkpoints trained with different seeds; for zero-shot methods, we use three different sampling seeds.
We report mean and standard deviation in both cases.
Empirically, we found the best performance with scales $S=2$, $\lambda_{\mathrm{HF}} = 0.2$, and $\lambda_{\text{aux}} = 0.01$.
For dataset construction, we employ \texttt{Qwen3-VL-8B}~\cite{bai2025qwen3vltechnicalreport} as VLM.

\noindent\textbf{Baselines.}
Due to high computational requirements, we focus our fine-tuning comparison on \texttt{Qwen-Image-Edit-2509}~\cite{qwen_image_edit} (in the following \texttt{Qwen-2509}) and \texttt{FLUX.1-Kontext\allowbreak-dev}~\cite{flux_1}.
We additionally report zero-shot performance of 
\texttt{FLUX.2-dev}~\cite{flux2}, 
\texttt{Kandinsky~5.0}~\cite{kanidsky5}, and 
\texttt{Qwen-Image-Edit-2511}~\cite{qwen_image_edit}.

\noindent\textbf{Evaluation Metrics.}
We evaluate generation quality using the Fréchet Inception Distance (\fid)~($\downarrow$)~\cite{fid}, structural fidelity and edge alignment with \ssim~($\uparrow$)~\cite{ssim} and Peak-Signal-to-Noise Ratio (\psnr~$\uparrow$), and perceptual similarity with \lpips~($\downarrow$)~\cite{lpips}.
For semantic alignment, we compute \clip~($\uparrow$)~\cite{clip} (\texttt{ViT-L/14@336px} ) and \dino~($\uparrow$)~\cite{dinov2} (\texttt{ViT-B/14-reg}) image-image similarity scores.
Metrics are computed using the TorchMetrics~\cite{torch_metrics} package when available.

\subsection{Can we leverage this space for supervision?}
\setcounter{rownum}{0}
\setlength{\tabcolsep}{2.4pt} %
\newcommand{\rownumber}{\stepcounter{rownum}\arabic{rownum}}
\begin{table*}[t!]
\centering
\small
\caption{Results on the \datasetName{} validation split. 
Indented rows denote addition of our auxiliary loss to each model.
We train each fine-tuned model 3 times with different seeds to report mean $\pm$ std (sampling seeds for zero-shot).
Best per section in \textbf{bold}.}
\label{table:val_results}
\sisetup{
    round-mode = places,
    round-precision = 3,
    detect-weight = true,
    mode = text,
    text-font-command = \footnotesize,
}
\begin{tabular}{@{}l
    S[table-format=2.2,round-precision=2] %
    S[table-format=0.3, round-precision=3,print-zero-integer=false]
    S[table-format=2.3, round-precision=2]
    S[table-format=0.3, round-precision=3,print-zero-integer=false]
    S[table-format=2.2, round-precision=2]
    S[table-format=2.2, round-precision=2]@{}}
\toprule
\textbf{Method} 
    & {\textbf{FID\,{\scriptsize$\downarrow$}}} 
    & {\textbf{SSIM\,{\scriptsize$\uparrow$}}} 
    & {\textbf{PSNR\,{\scriptsize$\uparrow$}}} 
    & {\textbf{LPIPS\,{\scriptsize$\downarrow$}}} 
    & {\textbf{CLIP\,{\scriptsize$\uparrow$}}} 
    & {\textbf{DINOv2\,{\scriptsize$\uparrow$}}} \\
\midrule
\multicolumn{7}{@{}l}{\textit{Zero-shot baselines}} \\[2pt]
\;\rownumber) Kandinsky~\cite{kanidsky5} & {\mstd{88.60}{1.15}} & {\mstd{.644}{.004}} & {\mstd{9.14}{.08}} & {\mstd{.579}{.007}} & {\mstd{83.55}{.38}} & {\mstd{56.74}{1.04}} \\
\;\rownumber) FLUX.1~\cite{flux_1}& {\mstd{59.32}{.39}} & {\mstd{.789}{.001}} & {\mstd{12.17}{.06}} & {\mstd{.345}{.003}} & {\mstd{90.76}{.10}} & {\mstd{78.14}{.25}} \\
\;\rownumber) FLUX.2~\cite{flux2} & {\mstd{\textbf{53.67}}{.30}} & {\mstd{.803}{.000}} & {\mstd{\textbf{12.80}}{.06}} & {\mstd{.320}{.000}} & {\mstd{\textbf{92.66}}{.00}} & {\mstd{\textbf{83.67}}{.25}} \\
\;\rownumber) Qwen-2511~\cite{qwen_image_edit} & {\mstd{64.23}{.47}} & {\mstd{.805}{.001}} & {\mstd{12.73}{.03}} & {\mstd{.336}{.000}} & {\mstd{89.17}{.22}} & {\mstd{77.83}{.31}} \\
\;\rownumber) Qwen-2509~\cite{qwen_image_edit}& {\mstd{57.66}{.02}} & {\mstd{\textbf{.818}}{.005}} & {\mstd{12.76}{.14}} & {\mstd{\textbf{.312}}{.007}} & {\mstd{91.95}{.07}} & {\mstd{82.30}{.06}} \\
\midrule
\multicolumn{7}{@{}l}{\textit{Fine-tuned on \datasetName{} (selected for best FID)}} \\[2pt]
\;\rownumber) FLUX.1~\cite{flux_1} 
    & {\mstd{31.68}{.18}} & {\mstd{.883}{.002}} 
    & {\mstd{18.62}{.05}} & {\mstd{.149}{.003}} 
    & {\mstd{96.20}{.10}} & {\mstd{90.99}{.37}} \\
\;\rownumber) \quad + \textit{Aux Loss}
    & {\mstd{\textbf{31.26}}{.18}} & {\mstd{\textbf{.883}}{.001}} 
    & {\mstd{\textbf{18.67}}{.05}} & {\mstd{\textbf{.148}}{.002}} 
    & {\mstd{\textbf{96.30}}{.10}} & {\mstd{\textbf{91.25}}{.13}} \\
\addlinespace[4pt] 
\;\rownumber) Qwen-2509~\cite{qwen_image_edit}
    & {\mstd{26.48}{.86}} & {\mstd{.893}{.001}}
    & {\mstd{19.64}{.03}} & {\mstd{.112}{.002}}
    & {\mstd{\textbf{97.21}}{.08}} & {\mstd{\textbf{93.98}}{.28}} \\
\;\rownumber) \quad + \textit{Aux Loss}
    & {\mstd{\textbf{25.91}}{.20}} & {\mstd{\textbf{.894}}{.001}} 
    & {\mstd{\textbf{19.79}}{.08}} & {\mstd{\textbf{.111}}{.001}} 
    & {\mstd{97.17}{.05}} & {\mstd{93.94}{.22}} \\

\bottomrule
\end{tabular}
\end{table*}

Table~\ref{table:val_results} summarizes all results on \datasetName. We first focus on zero-shot baselines (Rows 1–5), which highlight the inherent difficulty of the task.
Qualitative inspection reveals recurring failure modes, including incomplete object segmentation, fog-like artifacts, and difficulty disambiguating target products from visually similar products.
Rows~$6$--$9$ report results after fine-tuning on ABO-Edit, with checkpoints selected by validation FID over 3 separate training runs.

\noindent\textbf{Predictive Latent Space.}
We first evaluate \methodName on models that \textit{naturally} exhibit  predictive behavior in the conditioning embedding space (Observation 1).
As shown in Table~\ref{table:val_results}~(rows~$8$--$9$), fine-tuning with the proposed parameter-free auxiliary loss improves \fid ($26.48 \to \mathbf{25.91}$), alongside gains in \ssim ($.893 \to \mathbf{.894}$), \lpips ($.112\to\mathbf{.111}$), and \psnr ($19.64 \to \mathbf{19.79}$). 
These gains in pixel-level and perceptual similarity metrics indicate that our auxiliary loss encourages the model to better preserve fine-grained details and structural consistency.
We note a negligible decrease in \clip and \dino, indicating comparable semantic preservation.
Models trained with the auxiliary loss also exhibit lower variance across most metrics, suggesting more stable optimization dynamics.

\noindent\textbf{Non-Predictive Latent Space.}
We then evaluate FlowMirror on models that \textit{do not} exhibit predictive behavior in the conditioning space (Fig.~\ref{fig:image_internal_representation}).
As shown in Table~\ref{table:val_results}~(rows~$6$--$7$), the addition of the auxiliary loss consistently improves all metrics: \fid ($31.68 \to \mathbf{31.26}$), \ssim ($.883 \to \mathbf{.883}$), \lpips ($.149 \to \mathbf{.148}$), \psnr ($18.62 \to \mathbf{18.67}$), \clip ($96.20 \to \mathbf{96.30}$), and \dino ($90.99 \to \mathbf{91.25}$). This demonstrates that supervising the conditioning space is beneficial even when the model does not naturally encode target predictions, suggesting FlowMirror acts as an effective regularizer across rectified flow architectures.

\noindent\textbf{Human Evaluation.}
Our method was preferred 4.4 percentage points more often than the baseline (see \suppmat~\ref{supmat:human_eval} for details).

\subsection{Do Generative Models Have Spatial Awareness?}
We investigate whether editing models can accurately orient objects to specified viewpoints given precise rotation angles expressed in natural language.
We partition the validation set into three orientation buckets based on ground-truth target azimuth available in \datasetName: \textit{left}, \textit{front}, and \textit{right}.
For each generated image (using \datasetName validation prompts), we estimate the relative orientation with respect to the target using \texttt{Orient-Anything-v2}~\cite{orient_v2}. 
\setcounter{rownum}{0}
\begin{wraptable}[14]{r}{0.63\textwidth}
    \centering
    \small
    \caption{Azimuth error (°) between generated and target images on the validation set, bucketed by viewpoint. Lower is better. Results shown as mean {\scriptsize$\pm$ std}.}
    \label{tab:spatial_awareness}
    \begin{tabular}{@{}lccc@{}}
    \toprule
    \textbf{Model} & \textbf{Left} & \textbf{Front} & \textbf{Right} \\
    \midrule
    \multicolumn{4}{@{}l}{\textit{Zero-shot baselines}} \\[2pt]
    \;\rownumber) Kandinsky~\cite{kanidsky5}   & \mstd{47.2}{25.8} & \mstd{21.0}{22.0} & \mstd{37.2}{26.4} \\
    \;\rownumber) FLUX.1~\cite{flux_1}         & \mstd{21.7}{22.0} & \mstd{23.7}{17.7} & \mstd{38.9}{20.6} \\
    \;\rownumber) FLUX.2~\cite{flux2}          & \mstd{48.2}{25.2} & \mstd{11.9}{13.6} & \mstd{42.2}{21.0} \\
    \;\rownumber) Qwen-2511~\cite{qwen_image_edit} & \mstd{46.3}{26.5} & \mstd{18.7}{20.4} & \mstd{30.5}{17.2} \\
    \;\rownumber) Qwen-2509~\cite{qwen_image_edit} & \mstd{39.1}{28.0} & \mstd{18.5}{15.1} & \mstd{31.7}{17.2} \\
    \midrule
    \multicolumn{4}{@{}l}{\textit{Fine-tuned on \datasetName}} \\[2pt]
    \;\rownumber) \textbf{Qwen-2509} & \textbf{\mstd{2.4}{10.2}} & \textbf{\mstd{1.0}{2.2}} & \textbf{\mstd{1.2}{0.9}} \\
    \bottomrule
    \end{tabular}
\end{wraptable}

To ensure measurement reliability, we retain only samples with an ``exact'' predicted symmetry type (using \texttt{Orient-Anything-v2}), and discard those with azimuth errors exceeding \SI[round-mode=none]{90}{\degree}, which indicate orientation estimation failures.

\noindent\textbf{Results.}
Table~\ref{tab:spatial_awareness} reports the azimuth error distribution across viewpoints.
Fine-tuning on \datasetName{} yields substantial improvements in spatial control, achieving mean errors below \SI{2.5}{\degree}. This shows that task-specific training enables precise object orientation at degree-level accuracy without architectural changes.
In contrast, zero-shot models exhibit considerably larger errors (rows $1$--$5$), indicating that fine-grained angular placement from text prompts alone remains challenging without task-specific adaptation.
We also observe that some models show asymmetric performance across left and right views, which may reflect biases in their pre-training data distributions.

\noindent\textbf{Spatial Control.}
To demonstrate precise spatial control, we generate $5$ novel views at varying rotation angles.
As shown in Fig.~\ref{fig:sup:rotation_control}~(\suppmat), our fine-tuned model achieves fine-grained angular control while maintaining strong visual consistency across generated views.

\noindent\textbf{Failure Modes.}
As shown in Fig.~\ref{fig:sup:rotation_distributions_by_bucket}~(\suppmat), zero-shot models exhibit bimodal error distributions across viewpoint buckets.
This bimodality suggests two distinct failure modes:
\begin{enumerate*}[label=\textit{(\roman*)}]
    \item the model correctly interprets the target orientation but introduces 
    angular errors, and
    \item the model misinterprets the requested viewpoint entirely, rendering 
    an opposing orientation (\eg, right-facing instead of left-facing).
\end{enumerate*}
We provide per-model breakdowns with specific mode values in the supplementary 
material.
To ground these errors in concrete examples, we illustrate 
representative failure cases in 
Fig.~\ref{fig:sup:rotation_failures}~(\suppmat).

\subsection{Can Conditioning Embeddings Outpace Velocity Predictions?}
Fig.~\ref{fig:similarity_vs_step} (orange) reports the similarity between the conditioning embedding $\hCondL$ at timestep $t$ and the final generated image $\hatxzero$, computed by decoding $\hCondL$ with decoder $\mathcal{D}$ and measuring CLIP~\cite{clip} similarity.
Surprisingly, with our auxiliary loss, the model's predicted image in the conditioning space exhibits higher similarity to the target than the predicted velocity (green) at noisy timesteps, successfully encoding the target image.
At high noise ($t=1$), this gap is substantial (\textbf{$0.823$} \vs $0.723$).
Shape, structure, and overall quality are already encoded at early timesteps, leaving little room for the model to alter these properties.

\subsection{Analysis and Ablations}
As shown in Table~\ref{table:val_results} (rows 6--9), adding
$\mathcal{L}_{\text{aux}}$ improves generation quality.
Due to computational constraints, we now ablate 
its design choices on Qwen-Image-Edit-2509 using a single 
fixed seed.
We initially explored matching VAE channel-wise mean and std of
$\mathbf{h}_L^{\text{cond}}$ to $\mathbf{z}$, but this worsens \fid
($25.18\to26.15$). We attribute this to statistics conflicting with the spatially-grounded supervision of the loss.
Normalizing $\mathcal{L}_{\text{aux}}$ by the number of scales further helps ($26.52\to26.15$).
Reducing $\lambda_{\text{aux}}$ from 0.1 to 0.01 improves \fid ($25.93\to25.55$), indicating lighter regularization avoids interfering with the CFM objective.
Removing the HF term entirely ($\lambda_{\text{HF}}=0$) yields a high \fid among variants (27.07), confirming that high-frequency supervision is essential for detail preservation. However, over-weighting it ($\lambda_{\text{HF}}=1.0$) also degrades performance relative to $\lambda_{\text{HF}}=0.2$, as excessive detail pressure interferes with the LF structural signal.
Increasing beyond $S=2$ degrades performance, likely because finer pyramid levels lose meaningful structure through excessive downsampling.

\section{Conclusion \& Impact}
We presented three contributions.
First, \textit{ABO-Edit}, a benchmark of $12$k+ triplets pairing real-world product photographs with studio-quality targets 
rendered from artist-designed 3D assets, enabling controlled evaluation of geometry, texture, and fine-grained details.
Second, we revealed that rectified flow models encode target predictions in the conditioning embedding space $\hCondL$ even at high noise levels, and that these representations stabilize early in denoising.
Third, we proposed \textit{FlowMirror}, a parameter-free auxiliary loss that supervises this embedding space, improving generation quality across multiple metrics on two architectures.

\noindent\textbf{Limitations.}
A domain gap exists between our Blender-rendered targets and real photographs, particularly in material appearance and lighting. Our fine-tuning experiments were limited to DoRA adaptation with a $10$k-step budget; full fine-tuning may yield further gains. 
Finally, evaluation is restricted to \datasetName.

\noindent\textbf{Future Work.}
An important direction for future work is to close the synthetic--real gap via photorealistic rendering or using real image pairs, to expand to diverse product types, and to investigate \methodName's generalization to additional models and tasks.

\noindent\textbf{Impact.}
Our analysis opens new directions for future supervision strategies beyond CFM training, as demonstrated by \methodName's improvements.
\datasetName{} further enables research on fine-grained object preservation and degree-level viewpoint control (Fig.~\ref{fig:sup:rotation_control}, \suppmat), as well as the inverse task of generating lifestyle imagery from studio photographs.

\bibliographystyle{splncs04}
\bibliography{main}

@String(CVPR  = {IEEE Conf. Comput. Vis. Pattern Recog.})

@String(ICCV  = {Int. Conf. Comput. Vis.})

@String(NeurIPS = {Adv. Neural Inform. Process. Syst.})

@String(ICML  = {Int. Conf. Mach. Learn.})

@String(CVPR  = {CVPR})

@String(ICCV  = {ICCV})

@String(NeurIPS = {NeurIPS})

@String(ICML  = {ICML})

@inproceedings{abo_dataset, title={{ABO: Dataset and Benchmarks for Real-World 3D Object Understanding}}, url={http://dx.doi.org/10.1109/cvpr52688.2022.02045}, DOI={10.1109/cvpr52688.2022.02045}, booktitle={2022 IEEE/CVF Conference on Computer Vision and Pattern Recognition (CVPR)}, publisher={IEEE}, author={Collins, Jasmine and Goel, Shubham and Deng, Kenan and Luthra, Achleshwar and Xu, Leon and Gundogdu, Erhan and Zhang, Xi and Vicente, Tomas F. Yago and Dideriksen, Thomas and Arora, Himanshu and Guillaumin, Matthieu and Malik, Jitendra}, year={2022}, month=jun, pages={21094–21104} }

@misc{blender,
  title = {{Blender - a 3D Modelling and Rendering Package}},
  author = {{Blender Online Community}},
  organization = {Blender Foundation},
  year = {2025},
  url = {www.blender.org},
}

@misc{bai2025qwen3vltechnicalreport,
      title={{Qwen3-VL Technical Report}}, 
      author={Shuai Bai and Yuxuan Cai and Ruizhe Chen and Keqin Chen and Xionghui Chen and Zesen Cheng and Lianghao Deng and Wei Ding and Chang Gao and Chunjiang Ge and Wenbin Ge and Zhifang Guo and Qidong Huang and Jie Huang and Fei Huang and Binyuan Hui and Shutong Jiang and Zhaohai Li and Mingsheng Li and Mei Li and Kaixin Li and Zicheng Lin and Junyang Lin and Xuejing Liu and Jiawei Liu and Chenglong Liu and Yang Liu and Dayiheng Liu and Shixuan Liu and Dunjie Lu and Ruilin Luo and Chenxu Lv and Rui Men and Lingchen Meng and Xuancheng Ren and Xingzhang Ren and Sibo Song and Yuchong Sun and Jun Tang and Jianhong Tu and Jianqiang Wan and Peng Wang and Pengfei Wang and Qiuyue Wang and Yuxuan Wang and Tianbao Xie and Yiheng Xu and Haiyang Xu and Jin Xu and Zhibo Yang and Mingkun Yang and Jianxin Yang and An Yang and Bowen Yu and Fei Zhang and Hang Zhang and Xi Zhang and Bo Zheng and Humen Zhong and Jingren Zhou and Fan Zhou and Jing Zhou and Yuanzhi Zhu and Ke Zhu},
      year={2025},
      eprint={2511.21631},
      archivePrefix={arXiv},
      primaryClass={cs.CV},
      url={https://arxiv.org/abs/2511.21631}, 
}

@misc{yang2025qwen3technicalreport,
      title={{Qwen3 Technical Report}}, 
      author={An Yang and Anfeng Li and Baosong Yang and Beichen Zhang and Binyuan Hui and Bo Zheng and Bowen Yu and Chang Gao and Chengen Huang and Chenxu Lv and Chujie Zheng and Dayiheng Liu and Fan Zhou and Fei Huang and Feng Hu and Hao Ge and Haoran Wei and Huan Lin and Jialong Tang and Jian Yang and Jianhong Tu and Jianwei Zhang and Jianxin Yang and Jiaxi Yang and Jing Zhou and Jingren Zhou and Junyang Lin and Kai Dang and Keqin Bao and Kexin Yang and Le Yu and Lianghao Deng and Mei Li and Mingfeng Xue and Mingze Li and Pei Zhang and Peng Wang and Qin Zhu and Rui Men and Ruize Gao and Shixuan Liu and Shuang Luo and Tianhao Li and Tianyi Tang and Wenbiao Yin and Xingzhang Ren and Xinyu Wang and Xinyu Zhang and Xuancheng Ren and Yang Fan and Yang Su and Yichang Zhang and Yinger Zhang and Yu Wan and Yuqiong Liu and Zekun Wang and Zeyu Cui and Zhenru Zhang and Zhipeng Zhou and Zihan Qiu},
      year={2025},
      eprint={2505.09388},
      archivePrefix={arXiv},
      primaryClass={cs.CL},
      url={https://arxiv.org/abs/2505.09388}, 
}

@inproceedings{
    lipman2023flow,title={{Flow Matching for Generative Modeling}},author={Yaron Lipman and Ricky T. Q. Chen and Heli Ben-Hamu and Maximilian Nickel and Matthew Le},booktitle={The Eleventh International Conference on Learning Representations },year={2023},url={https://openreview.net/forum?id=PqvMRDCJT9t}
}

@inproceedings{
    flow_straight_and_fast,
    title={{Flow Straight and Fast: Learning to Generate and Transfer Data with Rectified Flow}},author={Xingchao Liu and Chengyue Gong and qiang liu},booktitle={The Eleventh International Conference on Learning Representations },
    year={2023},url={https://openreview.net/forum?id=XVjTT1nw5z}
}

@inproceedings{mmdit,
author = {Esser, Patrick and Kulal, Sumith and Blattmann, Andreas and Entezari, Rahim and M\"{u}ller, Jonas and Saini, Harry and Levi, Yam and Lorenz, Dominik and Sauer, Axel and Boesel, Frederic and Podell, Dustin and Dockhorn, Tim and English, Zion and Rombach, Robin},title = {{Scaling Rectified Flow Transformers for High-Resolution Image Synthesis}},
year = {2024},publisher = {JMLR.org},booktitle = {Proceedings of the 41st International Conference on Machine Learning},articleno = {503},numpages = {28},location = {Vienna, Austria},series = {ICML'24}, url={https://dl.acm.org/doi/10.5555/3692070.3692573}
}

@inproceedings{sdxl, title={{SDXL: Improving Latent Diffusion Models for High-Resolution Image Synthesis}},author={Dustin Podell and Zion English and Kyle Lacey and Andreas Blattmann and Tim Dockhorn and Jonas M{\"u}ller and Joe Penna and Robin Rombach},
booktitle={The Twelfth International Conference on Learning Representations},year={2024},url={https://openreview.net/forum?id=di52zR8xgf}
}

@INPROCEEDINGS{dit,
  author={Peebles, William and Xie, Saining},
  booktitle={2023 IEEE/CVF International Conference on Computer Vision (ICCV)}, 
  title={{Scalable Diffusion Models with Transformers}}, 
  year={2023},
  volume={},
  number={},
  pages={4172-4182},
  doi={10.1109/ICCV51070.2023.00387}}

@misc{flux_1,
      title={{FLUX.1 Kontext: Flow Matching for In-Context Image Generation and Editing in Latent Space}}, 
      author={Black Forest Labs and Stephen Batifol and Andreas Blattmann and Frederic Boesel and Saksham Consul and Cyril Diagne and Tim Dockhorn and Jack English and Zion English and Patrick Esser and Sumith Kulal and Kyle Lacey and Yam Levi and Cheng Li and Dominik Lorenz and Jonas Müller and Dustin Podell and Robin Rombach and Harry Saini and Axel Sauer and Luke Smith},
      year={2025},
      eprint={2506.15742},
      archivePrefix={arXiv},
      primaryClass={cs.GR},
      url={https://arxiv.org/abs/2506.15742}, 
}

@misc{qwen_image_edit,
      title={{Qwen-Image Technical Report}}, 
      author={Chenfei Wu and Jiahao Li and Jingren Zhou and Junyang Lin and Kaiyuan Gao and Kun Yan and Sheng-ming Yin and Shuai Bai and Xiao Xu and Yilei Chen and Yuxiang Chen and Zecheng Tang and Zekai Zhang and Zhengyi Wang and An Yang and Bowen Yu and Chen Cheng and Dayiheng Liu and Deqing Li and Hang Zhang and Hao Meng and Hu Wei and Jingyuan Ni and Kai Chen and Kuan Cao and Liang Peng and Lin Qu and Minggang Wu and Peng Wang and Shuting Yu and Tingkun Wen and Wensen Feng and Xiaoxiao Xu and Yi Wang and Yichang Zhang and Yongqiang Zhu and Yujia Wu and Yuxuan Cai and Zenan Liu},
      year={2025},
      eprint={2508.02324},
      archivePrefix={arXiv},
      primaryClass={cs.CV},
      url={https://arxiv.org/abs/2508.02324}, 
}

@InProceedings{sd1,
    author    = {Rombach, Robin and Blattmann, Andreas and Lorenz, Dominik and Esser, Patrick and Ommer, Bj\"orn},
    title     = {{High-Resolution Image Synthesis With Latent Diffusion Models}},
    booktitle = {Proceedings of the IEEE/CVF Conference on Computer Vision and Pattern Recognition (CVPR)},
    month     = {June},
    year      = {2022},
    pages     = {10684-10695},
    url={https://openaccess.thecvf.com/content/CVPR2022/html/Rombach_High-Resolution_Image_Synthesis_With_Latent_Diffusion_Models_CVPR_2022_paper}
}

@misc{flux2,
  title        = {{FLUX.2}},
  author       = {{Black Forest Labs}},
  year         = {2025},
  month        = jun,
  howpublished = {\url{https://bfl.ai/blog/flux-2}},
  note         = {Accessed: 2026}
}

@misc{qwen_image_lightning,
  author       = {ModelTC},
  title        = {Qwen-Image-Lightning},
  year         = {2024},
  publisher    = {GitHub},
  howpublished = {\url{https://github.com/ModelTC/Qwen-Image-Lightning}},
  note         = {Accessed: 2026}
}

@InProceedings{diffusion,
  title = 	 {{Deep Unsupervised Learning using Nonequilibrium Thermodynamics}},
  author = 	 {Sohl-Dickstein, Jascha and Weiss, Eric and Maheswaranathan, Niru and Ganguli, Surya},
  booktitle = 	 {Proceedings of the 32nd International Conference on Machine Learning},
  pages = 	 {2256--2265},
  year = 	 {2015},
  volume = 	 {37},
  series = 	 {Proceedings of Machine Learning Research},
  address = 	 {Lille, France},
  month = 	 {07--09 Jul},
  publisher =    {PMLR},
  url = 	 {https://proceedings.mlr.press/v37/sohl-dickstein15.html},
}

@InProceedings{clip,
  title = 	 {Learning Transferable Visual Models From Natural Language Supervision},
  author =       {Radford, Alec and Kim, Jong Wook and Hallacy, Chris and Ramesh, Aditya and Goh, Gabriel and Agarwal, Sandhini and Sastry, Girish and Askell, Amanda and Mishkin, Pamela and Clark, Jack and Krueger, Gretchen and Sutskever, Ilya},
  booktitle = 	 {Proceedings of the 38th International Conference on Machine Learning},
  pages = 	 {8748--8763},
  year = 	 {2021},
  volume = 	 {139},
  series = 	 {Proceedings of Machine Learning Research},
  month = 	 {18--24 Jul},
  publisher =    {PMLR},
  url = 	 {https://proceedings.mlr.press/v139/radford21a.html},
}

@article{li2025treft,
  title={{TReFT: Taming Rectified Flow Models For One-Step Image Translation}},
  author={Li, Shengqian and Gao, Ming and Liu, Yi and Lin, Zuzeng and Wang, Feng and Dai, Feng},
  journal={arXiv preprint arXiv:2511.20307},
  year={2025}
}

@inproceedings{fid,
 author = {Heusel, Martin and Ramsauer, Hubert and Unterthiner, Thomas and Nessler, Bernhard and Hochreiter, Sepp},
 booktitle = {Advances in Neural Information Processing Systems},
 pages = {},
 publisher = {Curran Associates, Inc.},
 title = {{GANs Trained by a Two Time-Scale Update Rule Converge to a Local Nash Equilibrium}},
 url = {https://proceedings.neurips.cc/paper_files/paper/2017/file/8a1d694707eb0fefe65871369074926d-Paper.pdf},
 volume = {30},
 year = {2017}
}

@ARTICLE{ssim,
  author={Zhou Wang and Bovik, A.C. and Sheikh, H.R. and Simoncelli, E.P.},
  journal={IEEE Transactions on Image Processing}, 
  title={{Image Quality Assessment: from Error Visibility to Structural Similarity}}, 
  year={2004},
  volume={13},
  number={4},
  pages={600-612},
  url={https://doi.org/10.1109/TIP.2003.819861},
  doi={10.1109/TIP.2003.819861}}

@inproceedings{ddpm,
 author = {Ho, Jonathan and Jain, Ajay and Abbeel, Pieter},
 booktitle = {Advances in Neural Information Processing Systems},
 pages = {6840--6851},
 publisher = {Curran Associates, Inc.},
 title = {{Denoising Diffusion Probabilistic Models}},
 url = {https://proceedings.neurips.cc/paper_files/paper/2020/file/4c5bcfec8584af0d967f1ab10179ca4b-Paper.pdf},
 volume = {33},
 year = {2020}
}

@inproceedings{video_diff_model,
 author = {Ho, Jonathan and Salimans, Tim and Gritsenko, Alexey and Chan, William and Norouzi, Mohammad and Fleet, David J},
 booktitle = {Advances in Neural Information Processing Systems},
 pages = {8633--8646},
 publisher = {Curran Associates, Inc.},
 title = {{Video Diffusion Models}},
 url = {https://proceedings.neurips.cc/paper_files/paper/2022/file/39235c56aef13fb05a6adc95eb9d8d66-Paper-Conference.pdf},
 volume = {35},
 year = {2022}
}

@misc{ho2022imagenvideohighdefinition,
      title={{Imagen Video: High Definition Video Generation with Diffusion Models}}, 
      author={Jonathan Ho and William Chan and Chitwan Saharia and Jay Whang and Ruiqi Gao and Alexey Gritsenko and Diederik P. Kingma and Ben Poole and Mohammad Norouzi and David J. Fleet and Tim Salimans},
      year={2022},
      eprint={2210.02303},
      archivePrefix={arXiv},
      primaryClass={cs.CV},
      url={https://arxiv.org/abs/2210.02303}, 
}

@misc{z_image,
      title={{Z-Image: An Efficient Image Generation Foundation Model with Single-Stream Diffusion Transformer}}, 
      author={Image Team and Huanqia Cai and Sihan Cao and Ruoyi Du and Peng Gao and Steven Hoi and Zhaohui Hou and Shijie Huang and Dengyang Jiang and Xin Jin and Liangchen Li and Zhen Li and Zhong-Yu Li and David Liu and Dongyang Liu and Junhan Shi and Qilong Wu and Feng Yu and Chi Zhang and Shifeng Zhang and Shilin Zhou},
      year={2025},
      eprint={2511.22699},
      archivePrefix={arXiv},
      primaryClass={cs.CV},
      url={https://arxiv.org/abs/2511.22699}, 
}

@misc{hiDream,
      title={{HiDream-I1: A High-Efficient Image Generative Foundation Model with Sparse Diffusion Transformer}}, 
      author={Qi Cai and Jingwen Chen and Yang Chen and Yehao Li and Fuchen Long and Yingwei Pan and Zhaofan Qiu and Yiheng Zhang and Fengbin Gao and Peihan Xu and Yimeng Wang and Kai Yu and Wenxuan Chen and Ziwei Feng and Zijian Gong and Jianzhuang Pan and Yi Peng and Rui Tian and Siyu Wang and Bo Zhao and Ting Yao and Tao Mei},
      year={2025},
      eprint={2505.22705},
      archivePrefix={arXiv},
      primaryClass={cs.CV},
      url={https://arxiv.org/abs/2505.22705}, 
}

@inproceedings{
in_context_edit,
title={{Enabling Instructional Image Editing with In-Context Generation in Large Scale Diffusion Transformer}},
author={Zechuan Zhang and Ji Xie and Yu Lu and Zongxin Yang and Yi Yang},
booktitle={The Thirty-ninth Annual Conference on Neural Information Processing Systems},
year={2025},
url={https://openreview.net/forum?id=5WyqKH9nOS}
}

@misc{ramesh2022hierarchicaltextconditionalimagegeneration,
      title={{Hierarchical Text-Conditional Image Generation with CLIP Latents}}, 
      author={Aditya Ramesh and Prafulla Dhariwal and Alex Nichol and Casey Chu and Mark Chen},
      year={2022},
      eprint={2204.06125},
      archivePrefix={arXiv},
      primaryClass={cs.CV},
      url={https://arxiv.org/abs/2204.06125}, 
}

@inproceedings{poole2023dreamfusion,
title={{DreamFusion: Text-to-3D using 2D Diffusion}},
author={Ben Poole and Ajay Jain and Jonathan T. Barron and Ben Mildenhall},
booktitle={The Eleventh International Conference on Learning Representations },
year={2023},
url={https://openreview.net/forum?id=FjNys5c7VyY}
}

@misc{wan2025wanopenadvancedlargescale,
      title={{Wan: Open and Advanced Large-Scale Video Generative Models}}, 
      author={Team Wan and Ang Wang and Baole Ai and Bin Wen and Chaojie Mao and Chen-Wei Xie and Di Chen and Feiwu Yu and Haiming Zhao and Jianxiao Yang and Jianyuan Zeng and Jiayu Wang and Jingfeng Zhang and Jingren Zhou and Jinkai Wang and Jixuan Chen and Kai Zhu and Kang Zhao and Keyu Yan and Lianghua Huang and Mengyang Feng and Ningyi Zhang and Pandeng Li and Pingyu Wu and Ruihang Chu and Ruili Feng and Shiwei Zhang and Siyang Sun and Tao Fang and Tianxing Wang and Tianyi Gui and Tingyu Weng and Tong Shen and Wei Lin and Wei Wang and Wei Wang and Wenmeng Zhou and Wente Wang and Wenting Shen and Wenyuan Yu and Xianzhong Shi and Xiaoming Huang and Xin Xu and Yan Kou and Yangyu Lv and Yifei Li and Yijing Liu and Yiming Wang and Yingya Zhang and Yitong Huang and Yong Li and You Wu and Yu Liu and Yulin Pan and Yun Zheng and Yuntao Hong and Yupeng Shi and Yutong Feng and Zeyinzi Jiang and Zhen Han and Zhi-Fan Wu and Ziyu Liu},
      year={2025},
      eprint={2503.20314},
      archivePrefix={arXiv},
      primaryClass={cs.CV},
      url={https://arxiv.org/abs/2503.20314}, 
}

@inproceedings{
    cfg,title={{Classifier-Free Diffusion Guidance}},author={Jonathan Ho and Tim Salimans},booktitle={NeurIPS 2021 Workshop on Deep Generative Models and Downstream Applications},year={2021},url={https://openreview.net/forum?id=qw8AKxfYbI}
}

@inproceedings{diffusion_design,
 author = {Karras, Tero and Aittala, Miika and Aila, Timo and Laine, Samuli},booktitle = {Advances in Neural Information Processing Systems}
,pages = {26565--26577},publisher = {Curran Associates, Inc.},title = {{Elucidating the Design Space of Diffusion-Based Generative Models}},url = {https://proceedings.neurips.cc/paper_files/paper/2022/file/a98846e9d9cc01cfb87eb694d946ce6b-Paper-Conference.pdf},volume = {35},year = {2022}
}

@inproceedings{ddim,title={{Denoising Diffusion Implicit Models}},author={Jiaming Song and Chenlin Meng and Stefano Ermon},booktitle={International Conference on Learning Representations},year={2021},url={https://openreview.net/forum?id=St1giarCHLP}
}

@article{t5,
  author  = {Colin Raffel and Noam Shazeer and Adam Roberts and Katherine Lee and Sharan Narang and Michael Matena and Yanqi Zhou and Wei Li and Peter J. Liu},
  title   = {{Exploring the Limits of Transfer Learning with a Unified Text-to-Text Transformer}},
  journal = {Journal of Machine Learning Research},
  year    = {2020},
  volume  = {21},
  number  = {140},
  pages   = {1--67},
  url     = {http://jmlr.org/papers/v21/20-074.html}
}

@misc{qwen_2_5_vl,
      title={{Qwen2.5-VL Technical Report}}, 
      author={Shuai Bai and Keqin Chen and Xuejing Liu and Jialin Wang and Wenbin Ge and Sibo Song and Kai Dang and Peng Wang and Shijie Wang and Jun Tang and Humen Zhong and Yuanzhi Zhu and Mingkun Yang and Zhaohai Li and Jianqiang Wan and Pengfei Wang and Wei Ding and Zheren Fu and Yiheng Xu and Jiabo Ye and Xi Zhang and Tianbao Xie and Zesen Cheng and Hang Zhang and Zhibo Yang and Haiyang Xu and Junyang Lin},
      year={2025},
      eprint={2502.13923},
      archivePrefix={arXiv},
      primaryClass={cs.CV},
      url={https://arxiv.org/abs/2502.13923}, 
}

@misc{nano_banana_pro,
  title        = {{Gemini 3 Pro Image Model Card}},
  author       = {{Google DeepMind}},
  year         = {2025},
  month        = {Nov},
  howpublished = {\url{https://storage.googleapis.com/deepmind-media/Model-Cards/Gemini-3-Pro-Image-Model-Card.pdf}},
  note         = {Model card published November 2025},
}

@misc{nano_banana,
  title        = {{Gemini 2.5 Flash Image Model Card}},
  author       = {{Google DeepMind}},
  year         = {2025},
  month        = {Sep},
  howpublished = {\url{https://storage.googleapis.com/deepmind-media/Model-Cards/Gemini-2-5-Flash-Model-Card.pdf}},
  note         = {Model card published/updated September 2025} ,
}

@InProceedings{lpips,
author = {Zhang, Richard and Isola, Phillip and Efros, Alexei A. and Shechtman, Eli and Wang, Oliver},
title = {{The Unreasonable Effectiveness of Deep Features as a Perceptual Metric}},
booktitle = {Proceedings of the IEEE Conference on Computer Vision and Pattern Recognition (CVPR)},
month = {June},
year = {2018},
url={https://openaccess.thecvf.com/content_cvpr_2018/html/Zhang_The_Unreasonable_Effectiveness_CVPR_2018_paper.html}
}

@article{
    dinov2,
    title={{DINOv2: Learning Robust Visual Features without Supervision}},
    author={Maxime Oquab and Timoth{\'e}e Darcet and Th{\'e}o Moutakanni and Huy V. Vo and Marc Szafraniec and Vasil Khalidov and Pierre Fernandez and Daniel HAZIZA and Francisco Massa and Alaaeldin El-Nouby and Mido Assran and Nicolas Ballas and Wojciech Galuba and Russell Howes and Po-Yao Huang and Shang-Wen Li and Ishan Misra and Michael Rabbat and Vasu Sharma and Gabriel Synnaeve and Hu Xu and Herve Jegou and Julien Mairal and Patrick Labatut and Armand Joulin and Piotr Bojanowski},
    journal={Transactions on Machine Learning Research},
    issn={2835-8856},
    year={2024},
    url={https://openreview.net/forum?id=a68SUt6zFt},
    note={Featured Certification}
}

@article{torch_metrics,
  url = {https://doi.org/10.21105/joss.04101},
  year = {2022},
  publisher = {The Open Journal},
  volume = {7},
  number = {70},
  pages = {4101},
  author = {Detlefsen, Nicki Skafte and Borovec, Jiri and Schock, Justus and Jha, Ananya Harsh and Koker, Teddy and Di Liello, Luca and Stancl, Daniel and Quan, Changsheng and Grechkin, Maxim and Falcon, William},
  title = {{TorchMetrics - Measuring Reproducibility in PyTorch}},
  journal = {Journal of Open Source Software}
}

@inproceedings{
    liu2024dora,
    title={{DoRA: Weight-Decomposed Low-Rank Adaptation}},
    author={Shih-yang Liu and Chien-Yi Wang and Hongxu Yin and Pavlo Molchanov and Yu-Chiang Frank Wang and Kwang-Ting Cheng and Min-Hung Chen},
    booktitle={Forty-first International Conference on Machine Learning},
    year={2024},
    url={https://openreview.net/forum?id=3d5CIRG1n2}
}

@inproceedings{
    orient_v2,
    title={{Orient Anything V2: Unifying Orientation and Rotation Understanding}},
    author={Zehan Wang and Ziang Zhang and Jiayang Xu and Jialei Wang and Tianyu Pang and Chao Du and Hengshuang Zhao and Zhou Zhao},
    booktitle={The Thirty-ninth Annual Conference on Neural Information Processing Systems},
    year={2025},
    url={https://openreview.net/forum?id=n3armuTFit}
}

@misc{kanidsky5,
      title={{Kandinsky 5.0: A Family of Foundation Models for Image and Video Generation}}, 
      author={Vladimir Arkhipkin and Vladimir Korviakov and Nikolai Gerasimenko and Denis Parkhomenko and Viacheslav Vasilev and Alexey Letunovskiy and Nikolai Vaulin and Maria Kovaleva and Ivan Kirillov and Lev Novitskiy and Denis Koposov and Nikita Kiselev and Alexander Varlamov and Dmitrii Mikhailov and Vladimir Polovnikov and Andrey Shutkin and Julia Agafonova and Ilya Vasiliev and Anastasiia Kargapoltseva and Anna Dmitrienko and Anastasia Maltseva and Anna Averchenkova and Olga Kim and Tatiana Nikulina and Denis Dimitrov},
      year={2025},
      eprint={2511.14993},
      archivePrefix={arXiv},
      primaryClass={cs.CV},
      url={https://arxiv.org/abs/2511.14993}, 
}

@InProceedings{Wang_2025_CVPR,
    author    = {Wang, Zhendong and Bao, Jianmin and Gu, Shuyang and Chen, Dong and Zhou, Wengang and Li, Houqiang},
    title     = {{DesignDiffusion: High-Quality Text-to-Design Image Generation with Diffusion Models}},
    booktitle = {Proceedings of the IEEE/CVF Conference on Computer Vision and Pattern Recognition (CVPR)},
    month     = {June},
    year      = {2025},
    pages     = {20906-20915},
url={https://openaccess.thecvf.com/content/CVPR2025/html/Wang_DesignDiffusion_High-Quality_Text-to-Design_Image_Generation_with_Diffusion_Models_CVPR_2025_paper.html}
}

@InProceedings{Lin_2023_CVPR,
    author    = {Lin, Chen-Hsuan and Gao, Jun and Tang, Luming and Takikawa, Towaki and Zeng, Xiaohui and Huang, Xun and Kreis, Karsten and Fidler, Sanja and Liu, Ming-Yu and Lin, Tsung-Yi},
    title     = {{Magic3D: High-Resolution Text-to-3D Content Creation}},
    booktitle = {Proceedings of the IEEE/CVF Conference on Computer Vision and Pattern Recognition (CVPR)},
    month     = {June},
    year      = {2023},
    pages     = {300-309},
url={https://openaccess.thecvf.com/content/CVPR2023/html/Lin_Magic3D_High-Resolution_Text-to-3D_Content_Creation_CVPR_2023_paper.html}
}

@InProceedings{Cao_2023_ICCV,
    author    = {Cao, Mingdeng and Wang, Xintao and Qi, Zhongang and Shan, Ying and Qie, Xiaohu and Zheng, Yinqiang},
    title     = {{MasaCtrl: Tuning-Free Mutual Self-Attention Control for Consistent Image Synthesis and Editing}},
    booktitle = {Proceedings of the IEEE/CVF International Conference on Computer Vision (ICCV)},
    month     = {October},
    year      = {2023},
    pages     = {22560-22570},
    url={https://openaccess.thecvf.com/content/ICCV2023/html/Cao_MasaCtrl_Tuning-Free_Mutual_Self-Attention_Control_for_Consistent_Image_Synthesis_and_ICCV_2023_paper.html}
}

@inproceedings{
hertz2023prompttoprompt,
title={{Prompt-to-Prompt Image Editing with Cross-Attention Control}},
author={Amir Hertz and Ron Mokady and Jay Tenenbaum and Kfir Aberman and Yael Pritch and Daniel Cohen-Or},
booktitle={The Eleventh International Conference on Learning Representations },
year={2023},
url={https://openreview.net/forum?id=_CDixzkzeyb}
}

@INPROCEEDINGS {1_to_3d,
author = { Ye, Jianglong and Wang, Peng and Li, Kejie and Shi, Yichun and Wang, Heng },
booktitle = { 2024 International Conference on 3D Vision (3DV) },
title = {{ Consistent-1-to-3: Consistent Image to 3D View Synthesis via Geometry-aware Diffusion Models }},
year = {2024},
volume = {},
ISSN = {},
pages = {664-674},
doi = {10.1109/3DV62453.2024.00027},
url = {https://doi.ieeecomputersociety.org/10.1109/3DV62453.2024.00027},
publisher = {IEEE Computer Society},
address = {Los Alamitos, CA, USA},
month =mar}

@InProceedings{Yu_2023_ICCV,
    author    = {Yu, Jason J. and Forghani, Fereshteh and Derpanis, Konstantinos G. and Brubaker, Marcus A.},
    title     = {{Long-Term Photometric Consistent Novel View Synthesis with Diffusion Models}},
    booktitle = {Proceedings of the IEEE/CVF International Conference on Computer Vision (ICCV)},
    month     = {October},
    year      = {2023},
    pages     = {7094-7104},
url={https://openaccess.thecvf.com/content/ICCV2023/html/Yu_Long-Term_Photometric_Consistent_Novel_View_Synthesis_with_Diffusion_Models_ICCV_2023_paper.html}
}

@article{Brylla01102020, title={{Scene Sells: Why Spatial Backgrounds Outperform Isolated Product Depictions Online}}, volume={24}, ISSN={1557-9301}, url={http://dx.doi.org/10.1080/10864415.2020.1806470}, DOI={10.1080/10864415.2020.1806470}, number={4}, journal={International Journal of Electronic Commerce}, publisher={Informa UK Limited}, author={Brylla, Daniel and Walsh, Gianfranco}, year={2020}, month=oct, pages={497–526} }

@article{Maier02012019, title={{The Negative Effect of Product Image Inconsistency on Product Overviews During the Online Product Search}}, volume={23}, ISSN={1557-9301}, url={http://dx.doi.org/10.1080/10864415.2018.1512281}, DOI={10.1080/10864415.2018.1512281}, number={1}, journal={International Journal of Electronic Commerce}, publisher={Informa UK Limited}, author={Maier, Erik}, year={2019}, month=jan, pages={110–143} }

@article{diffusion_survey,
author = {Yang, Ling and Zhang, Zhilong and Song, Yang and Hong, Shenda and Xu, Runsheng and Zhao, Yue and Zhang, Wentao and Cui, Bin and Yang, Ming-Hsuan},
title = {{Diffusion Models: A Comprehensive Survey of Methods and Applications}},
year = {2023},
issue_date = {April 2024},
publisher = {Association for Computing Machinery},
address = {New York, NY, USA},
volume = {56},
number = {4},
issn = {0360-0300},
url = {https://doi.org/10.1145/3626235},
doi = {10.1145/3626235},
journal = {ACM Comput. Surv.},
month = nov,
articleno = {105},
numpages = {39}
}

@InProceedings{edit_bench,
    author    = {Wang, Su and Saharia, Chitwan and Montgomery, Ceslee and Pont-Tuset, Jordi and Noy, Shai and Pellegrini, Stefano and Onoe, Yasumasa and Laszlo, Sarah and Fleet, David J. and Soricut, Radu and Baldridge, Jason and Norouzi, Mohammad and Anderson, Peter and Chan, William},
    title     = {{Imagen Editor and EditBench: Advancing and Evaluating Text-Guided Image Inpainting}},
    booktitle = {Proceedings of the IEEE/CVF Conference on Computer Vision and Pattern Recognition (CVPR)},
    month     = {June},
    year      = {2023},
    pages     = {18359-18369},
    url={https://openaccess.thecvf.com/content/CVPR2023/html/Wang_Imagen_Editor_and_EditBench_Advancing_and_Evaluating_Text-Guided_Image_Inpainting_CVPR_2023_paper.html}
}

@inproceedings{magic_brush,
 author = {Zhang, Kai and Mo, Lingbo and Chen, Wenhu and Sun, Huan and Su, Yu},
 booktitle = {Advances in Neural Information Processing Systems},
 pages = {31428--31449},
 publisher = {Curran Associates, Inc.},
 title = {{MagicBrush: A Manually Annotated Dataset for Instruction-Guided Image Editing}},
 url = {https://proceedings.neurips.cc/paper_files/paper/2023/file/64008fa30cba9b4d1ab1bd3bd3d57d61-Paper-Datasets_and_Benchmarks.pdf},
 volume = {36},
 year = {2023}
}

@InProceedings{instruct_pix2pix,
    author    = {Brooks, Tim and Holynski, Aleksander and Efros, Alexei A.},
    title     = {{InstructPix2Pix: Learning To Follow Image Editing Instructions}},
    booktitle = {Proceedings of the IEEE/CVF Conference on Computer Vision and Pattern Recognition (CVPR)},
    month     = {June},
    year      = {2023},
    pages     = {18392-18402},
    url={https://openaccess.thecvf.com/content/CVPR2023/html/Brooks_InstructPix2Pix_Learning_To_Follow_Image_Editing_Instructions_CVPR_2023_paper.html}
}

@inproceedings{automati_gen,
author = {Fan, Xiaochuan and Zhang, Chi and Yang, Yong and Shang, Yue and Zhang, Xueying and He, Zhen and Xiao, Yun and Long, Bo and Wu, Lingfei},
title = {{Automatic Generation of Product-Image Sequence in E-commerce}},
year = {2022},
isbn = {9781450393850},
publisher = {Association for Computing Machinery},
address = {New York, NY, USA},
url = {https://doi.org/10.1145/3534678.3539149},
doi = {10.1145/3534678.3539149},
booktitle = {Proceedings of the 28th ACM SIGKDD Conference on Knowledge Discovery and Data Mining},
pages = {2851–2859},
numpages = {9},
location = {Washington DC, USA},
series = {KDD '22}
}

@inproceedings{thousand_word,
author = {Di, Wei and Sundaresan, Neel and Piramuthu, Robinson and Bhardwaj, Anurag},
title = {{Is a Picture Really Worth a Thousand Words? - On the Role of images in E-commerce}},
year = {2014},
isbn = {9781450323512},
publisher = {Association for Computing Machinery},
address = {New York, NY, USA},
url = {https://doi.org/10.1145/2556195.2556226},
doi = {10.1145/2556195.2556226},
booktitle = {Proceedings of the 7th ACM International Conference on Web Search and Data Mining},
pages = {633–642},
numpages = {10},
location = {New York, New York, USA},
series = {WSDM '14}
}

@InProceedings{Sheynin_2024_CVPR,
    author    = {Sheynin, Shelly and Polyak, Adam and Singer, Uriel and Kirstain, Yuval and Zohar, Amit and Ashual, Oron and Parikh, Devi and Taigman, Yaniv},
    title     = {{Emu Edit: Precise Image Editing via Recognition and Generation Tasks}},
    booktitle = {Proceedings of the IEEE/CVF Conference on Computer Vision and Pattern Recognition (CVPR)},
    month     = {June},
    year      = {2024},
    pages     = {8871-8879},
    url ={https://openaccess.thecvf.com/content/CVPR2024/html/Sheynin_Emu_Edit_Precise_Image_Editing_via_Recognition_and_Generation_Tasks_CVPR_2024_paper.html}
}

@misc{picobanana,
      title={{Pico-Banana-400K: A Large-Scale Dataset for Text-Guided Image Editing}}, 
      author={Yusu Qian and Eli Bocek-Rivele and Liangchen Song and Jialing Tong and Yinfei Yang and Jiasen Lu and Wenze Hu and Zhe Gan},
      year={2025},
      eprint={2510.19808},
      archivePrefix={arXiv},
      primaryClass={cs.CV},
      url={https://arxiv.org/abs/2510.19808}, 
}

@inproceedings{
ye2025imgedit,
title={{ImgEdit: A Unified Image Editing Dataset and Benchmark}},
author={Yang Ye and Xianyi He and Zongjian Li and Bin Lin and Shenghai Yuan and Zhiyuan Yan and Bohan Hou and Li Yuan},
booktitle={The Thirty-ninth Annual Conference on Neural Information Processing Systems Datasets and Benchmarks Track},
year={2025},
url={https://openreview.net/forum?id=uUCSrMlfD3}
}

@inproceedings{
hui2025hqedit,
title={{HQ-Edit: A High-Quality Dataset for Instruction-based Image Editing}},
author={Mude Hui and Siwei Yang and Bingchen Zhao and Yichun Shi and Heng Wang and Peng Wang and Cihang Xie and Yuyin Zhou},
booktitle={The Thirteenth International Conference on Learning Representations},
year={2025},
url={https://openreview.net/forum?id=mZptYYttFj}
}

@inproceedings{NEURIPS2024_05a30a0f,
 author = {Zhao, Haozhe and Ma, Xiaojian and Chen, Liang and Si, Shuzheng and Wu, Rujie and An, Kaikai and Yu, Peiyu and Zhang, Minjia and Li, Qing and Chang, Baobao},
 booktitle = {Advances in Neural Information Processing Systems},
 doi = {10.52202/079017-0100},
 pages = {3058--3093},
 publisher = {Curran Associates, Inc.},
 title = {{UltraEdit: Instruction-based Fine-Grained Image Editing at Scale}},
 url = {https://proceedings.neurips.cc/paper_files/paper/2024/file/05a30a0fc9e6bacdd3abd4ca8508a9e6-Paper-Datasets_and_Benchmarks_Track.pdf},
 volume = {37},
 year = {2024}
}

@misc{human_edit,
      title={{HumanEdit: A High-Quality Human-Rewarded Dataset for Instruction-based Image Editing}}, 
      author={Jinbin Bai and Wei Chow and Ling Yang and Xiangtai Li and Juncheng Li and Hanwang Zhang and Shuicheng Yan},
      year={2025},
      eprint={2412.04280},
      archivePrefix={arXiv},
      primaryClass={cs.CV},
      url={https://arxiv.org/abs/2412.04280}, 
}

% \makeatletter
% \renewcommand{\protected@write}[3]{%
%   \begingroup
%   #2%
%   \let\protect\@unexpandable@protect
%   \edef\reserved@a{\immediate\write#1{#3}}%
%   \reserved@a
%   \endgroup}
% \makeatother
% \AtBeginShipout{\AtBeginShipoutDiscard}
\section{Appendix}
\label{section:supplementary}
\setcounter{section}{7}
In this appendix, we provide additional details organized as follows:
\begin{itemize}
    \item \textbf{\datasetName dataset} (\underline{Sec.~\ref{supmat:ourdataset}}): 
    we present the static prompt used for dataset construction~(Sec.~\ref{sup:dataset:prompt:static_prompt_dataset}), 
    analyze product type distribution~(Sec.~\ref{sup:dataset:product_types_distribution}),
    describe the angle distribution during target image rendering~(Sec.~\ref{sup:dataset:rotation_examples}), 
    and report image resolution statistics~(Sec.~\ref{sup:dataset:image_resolution_statistic}).
    Finally, we provide details on human quality control (Sec.~\ref{sup:dataset:human_quality_control}) and provide a discussion of related datasets~(Sec.~\ref{sup:dataset:related_work_dataset}).

    \item \textbf{Spatial Awareness}~(\underline{Sec.~\ref{sup:sec:spatial_awareness}}): we first show the fine-grained control and spatial understanding of our method (Sec.~\ref{sup:spatial_Exp:fine_grained_control}). Then, we analyze the distribution of azimuth error between generated and target images across fine-tuned and zero-shot models~(Sec.~\ref{ssup:spatial_Exp:azimuth}),
    and visualize examples of such failures~(Sec.~\ref{ssup:spatial_Exp:failures}).
    \item \textbf{Human Evaluation}~(\underline{Sec.~\ref{supmat:human_eval}}): we report details regarding the human evaluation for the blind pairwise comparison between our method and the baseline.
    \item \textbf{Flow Matching}~(\underline{Sec.~\ref{supmat:flow_model}}): we derive the $\hatxzero$ estimation used in our formulation.

\end{itemize}

\section{\datasetName}
\label{supmat:ourdataset}
\subsection{Prompt for Dataset Construction.}
\label{sup:dataset:prompt:static_prompt_dataset}
Template Prompt (dataset generation):
\begin{tcolorbox}[colback=gray!10, colframe=gray!50, 
    fontupper=\small\ttfamily, breakable]
Generate an image of the \{OBJECT\_DESC\} on a pure white background.\\ 
Ensure strict preservation of the \{OBJECT\_CATEGORY\}'s precise geometry, edges, intersections, and proportions. \\
Remove people, info-graphics, and text.\\
The entire \{OBJECT\_CATEGORY\} is fully visible (not cropped) and isolated, no other objects or text in the scene.\\
Ensure a soft natural lighting with gentle shadows.\\
\{ROTATION\_INFORMATION\}. The rotation coordinate system is the front-view of the object.
\end{tcolorbox}

\subsection{Product Types Distribution.}
\label{sup:dataset:product_types_distribution}
\begin{figure*}[t]
  \centering
  \begin{subfigure}{0.48\linewidth}
    \centering
    \includegraphics[width=\linewidth, trim={0 0 0.3cm 1.5cm}, clip]{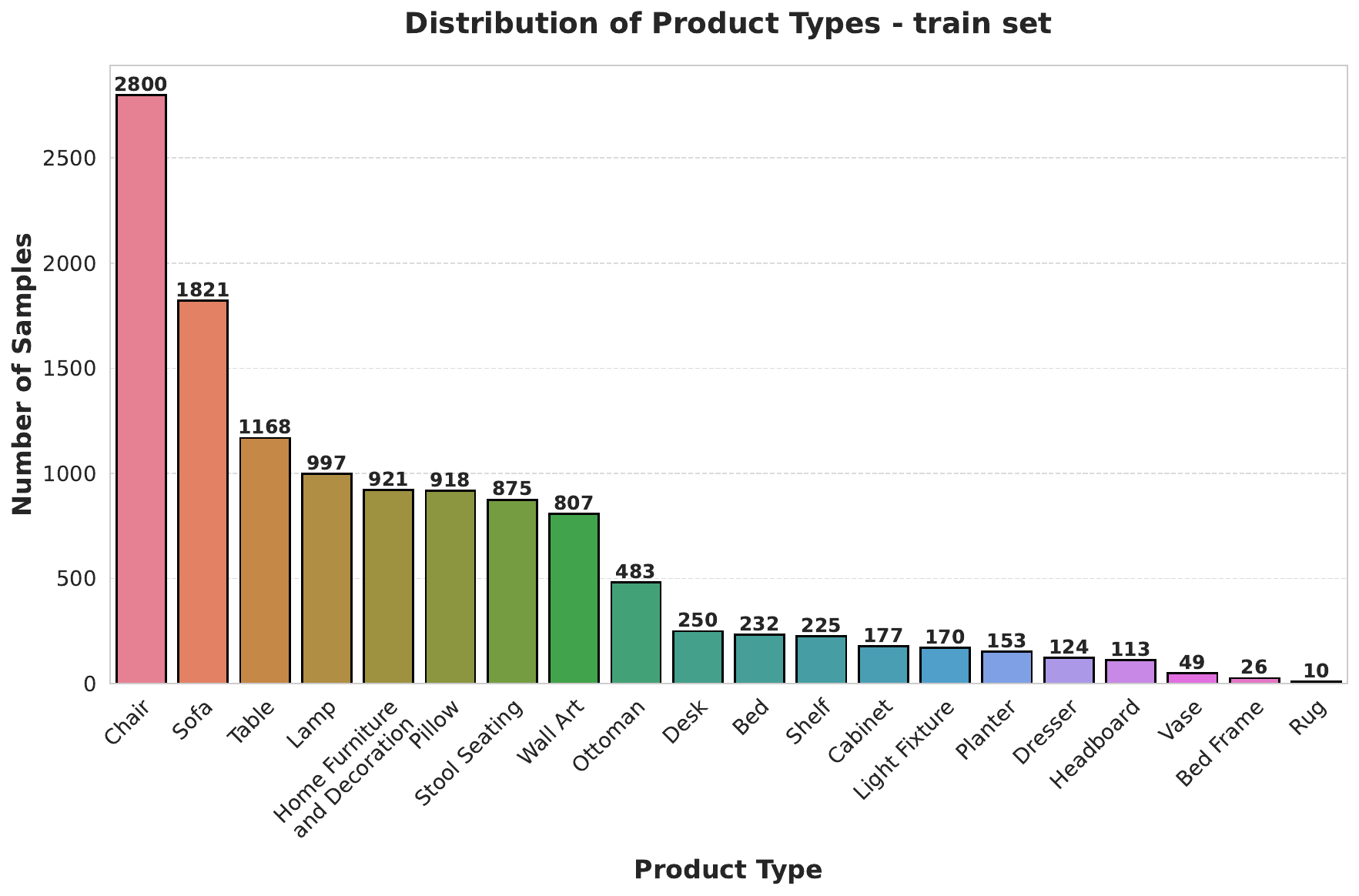}
    \caption{Training set.}
    \label{fig:sup:dataset:product_type_distribution_train}
  \end{subfigure}
  \hfill
  \begin{subfigure}{0.48\linewidth}
    \centering
    \includegraphics[width=\linewidth, trim={0 0 0.3cm 1.5cm}, clip]{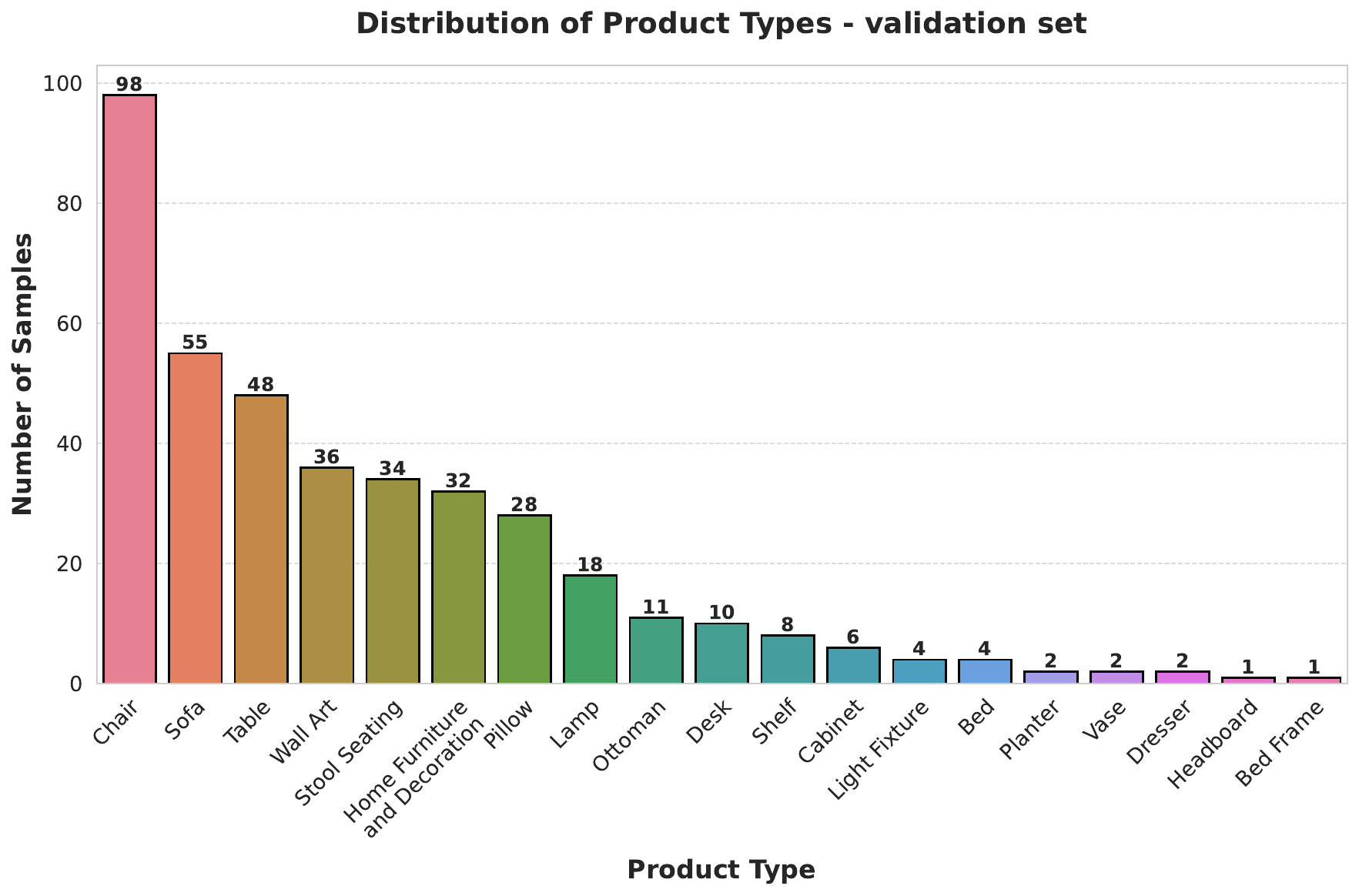}
    \caption{Validation set.}
    \label{fig:sup:dataset:product_type_distribution_validation}
  \end{subfigure}
  \caption{Distribution of product types across dataset splits.}
  \label{fig:sup:product_distribution}
\end{figure*}

Fig.~\ref{fig:sup:product_distribution} presents the distribution of product types across dataset splits. The training set (Fig.~\ref{fig:sup:dataset:product_type_distribution_train}) contains $12{,}319$ images, while the validation set (Fig.~\ref{fig:sup:dataset:product_type_distribution_validation}) comprises $400$ samples. 
Both splits exhibit a long-tailed distribution, with certain product categories being significantly more prevalent than others.

\subsection{Rotation Examples.}
\label{sup:dataset:rotation_examples}
\begin{figure*}[h]
  \centering
  \begin{subfigure}{0.5\linewidth}
      \centering
      \includegraphics[width=\linewidth]{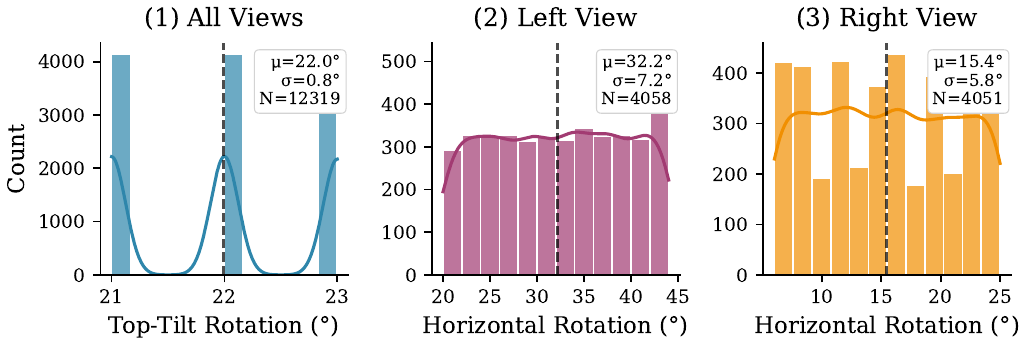}
      \caption{Rotation angle distribution (\textit{training} set).}
      \label{fig:sup:dataset:rotation_train}
  \end{subfigure}%
  \hfill  %
  \begin{subfigure}{0.5\linewidth}
    \centering
    \includegraphics[trim={0.5cm 0cm 0 0}, clip, width=\linewidth]{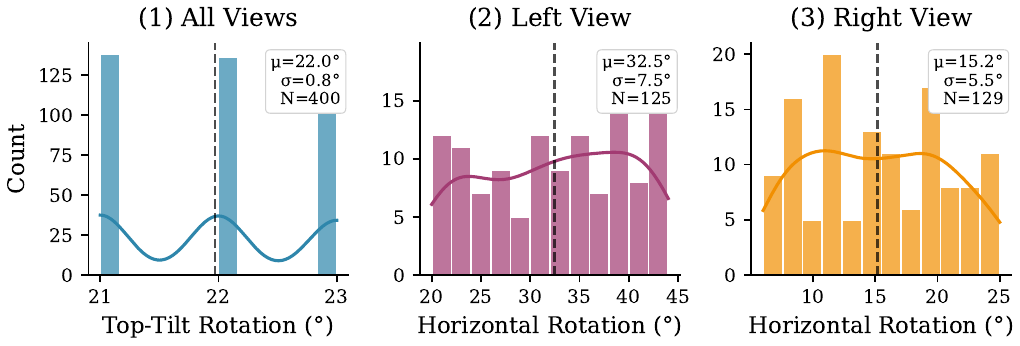}
    \caption{Rotation angle distribution (\textit{validation} set).}
    \label{fig:sup:dataset:rotation_val}
  \end{subfigure}
  \caption{Distribution of rotation angles across dataset splits (mean $\pm$ std), bucketed by object rotation. 
  (\textbf{a}-1, \textbf{b}-1) show the tilt angle distribution across all views.
  (\textbf{a}-2/3, \textbf{b}-2/3) show the azimuth angle distribution for the \textit{left} and \textit{right} views specifically. The \textit{front} view is omitted as its azimuth is \SI{0}{\degree} by definition.}
  \label{fig:sup:rotation_distribution}
\end{figure*}

In Fig.~\ref{fig:sup:rotation_distribution}, we show the rotation distribution of the train and validation set.
We report the distribution of rotation angles across dataset splits (mean $\pm$ std), bucketed by object rotation. 

\begin{figure*}[h]
  \centering
  \includegraphics[width=0.9\linewidth]{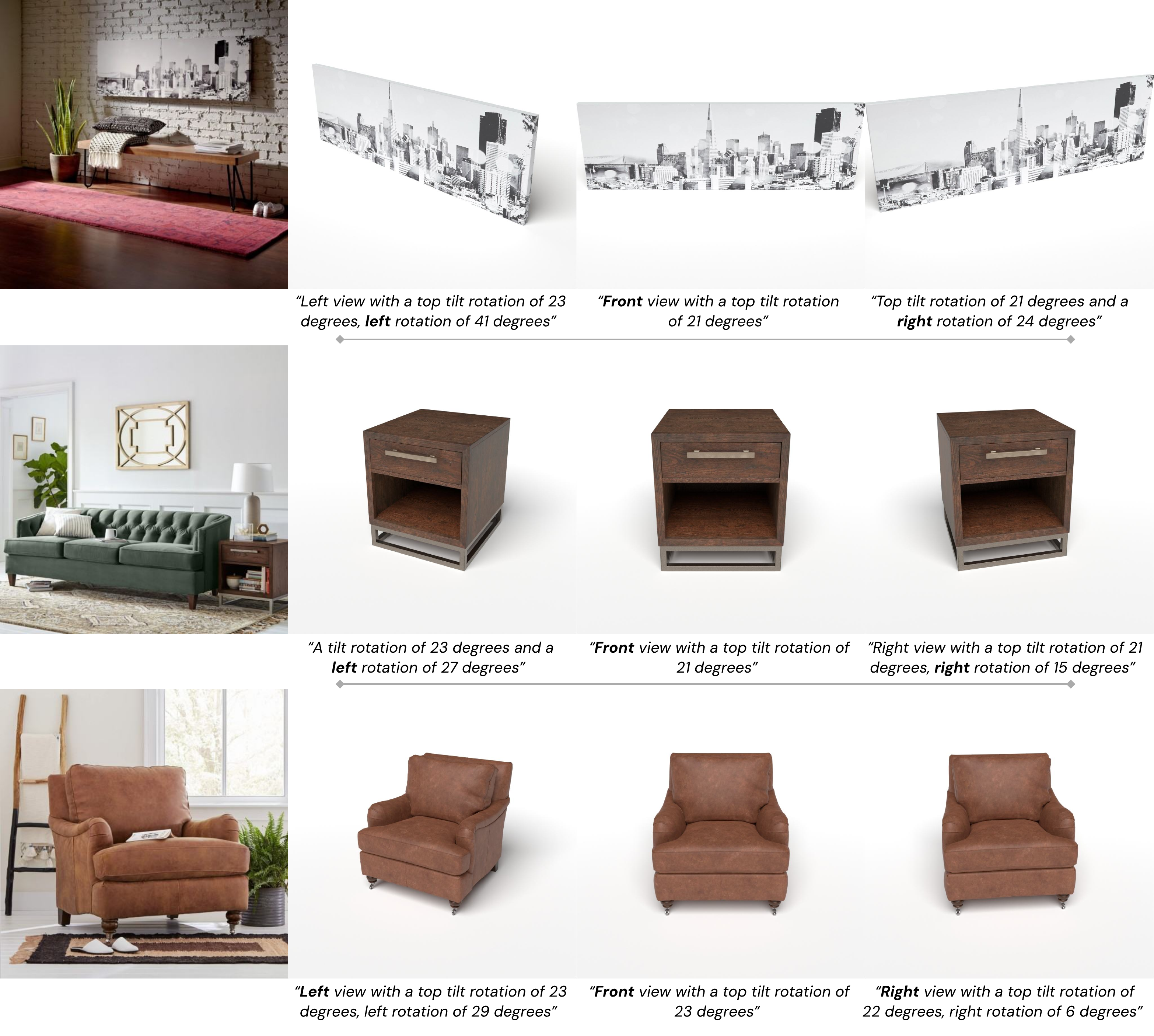}
  \caption{We show representative samples for each of the three views: \textit{front}, \textit{left}, and \textit{right}. For each sample, we also display the rotation parameters, specifying the target view and the angle by which the object is rotated.}
  \label{fig:sup:rotation_information}
\end{figure*}

We first report the tilt angle across all views (\textit{left}, \textit{front}, and \textit{right}), with a mean tilt rotation of \SI{22}{\degree} (\textbf{a}-1, \textbf{b}-1).
Then, we report the azimuth angle for the \textit{left} and \textit{right} views (\textbf{a}-2/3, \textbf{b}-2/3); the \textit{front} view has a fixed azimuth of \SI{0}{\degree}.
For the \textit{left} view, we sample angles uniformly between \SI{20}{\degree} and \SI{45}{\degree}, mimicking conventional e-commerce product placement.
For the \textit{right} view, we sample angles between \SI{5}{\degree} and \SI{25}{\degree}, representing a narrower range.
Furthermore, Fig.~\ref{fig:sup:rotation_information} shows representative samples from our proposed \datasetName.
The source images depict challenging scenarios with products in use, while the corresponding renders illustrate high-quality outputs across three distinct viewpoints per object.
\subsection{Image Resolution Statistics.}
\label{sup:dataset:image_resolution_statistic}
\begin{figure*}[h]
  \centering
  \includegraphics[width=0.4\linewidth]{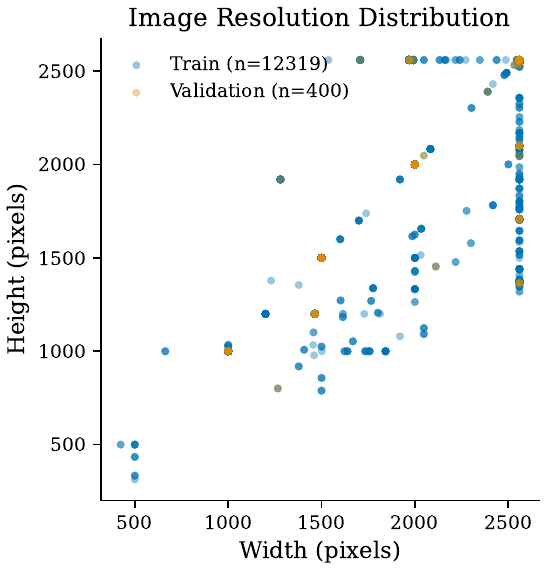}
    \caption{Source images resolution distribution across train (blue) and validation (orange) splits.}
  \label{fig:sup:resolutions_distributions}
\end{figure*}
\datasetName comprises high-resolution images suitable for fine-grained visual analysis, as shown in Fig.~\ref{fig:sup:resolutions_distributions}.
The training set contains  12,319 source images with a mean resolution of $2361 \times 2338$ pixels (width $\times$ height), while the validation set contains 400 source images with comparable statistics (mean: $2373 \times 2353$ pixels). 
The high average resolution enables detailed object-level analysis and supports downstream tasks requiring fine-grained visual features.
\subsection{Human Quality Control.}
\label{sup:dataset:human_quality_control}
First, to minimize annotator cognitive load, we use an LLM (\texttt{Qwen3-4B}~\cite{yang2025qwen3technicalreport}) to shorten the prompts $\pStoT$ by removing unnecessary details while retaining the core information, \ie, object description, background specification, and rotation parameters.
We then define the following rejection criteria:
\begin{enumerate*}[label=\textit{(\roman*)}]
\item \textit{Object Visibility}:~the object in the source image is partially cropped or occluded;
\item \textit{Geometry Artifacts}:~the object's shape, structure, or proportions are not preserved in the target image;
\item \textit{Prompt Mismatch}:~the rotation of the product in the target image does not match the rotation specified in the prompt;
\item \textit{Appearance Artifacts}:~material details, fabric patterns, or colors are inconsistent between source and target, or the rendered object exhibits unnatural shadows or incorrect exposure; and
\item \textit{Hallucinated Object}:~the rendered product does not correspond to the product depicted in the source image.
\end{enumerate*}
We proceed by designing an HTML annotation template containing the task guidelines along with representative examples of accepted and rejected triplets. The template was iteratively refined over several pilot rounds to improve annotator consensus. 
After finalizing the guidelines, we perform full-scale annotation.

To evaluate annotation quality, we manually label a reference set of 500 triplets, which we treat as ground truth. 
We compare several label aggregation strategies against this set and find that majority vote yields the highest agreement, achieving high precision and recall on this held-out validation set.

\subsection{Related Works: Datasets.}\label{sup:dataset:related_work_dataset}
Several benchmarks have been proposed for text-guided image editing, varying in scale, annotation methodology, and task coverage. EditBench~\cite{edit_bench} introduced a text-guided image inpainting benchmark comprising $240$ images with masked-region annotations. InstructPix2Pix~\cite{instruct_pix2pix} pioneered large-scale synthetic editing pair generation by combining GPT-3 and Prompt-to-Prompt~\cite{hertz2023prompttoprompt}. HQ-Edit~\cite{hui2025hqedit} proposed a data collection pipeline that leverages advanced foundation models to improve editing quality. MagicBrush~\cite{magic_brush} provides $10$K human-annotated triplets of source images, editing instructions, and target images. Similarly, Emu-Edit~\cite{Sheynin_2024_CVPR} proposes a benchmark that includes seven different image editing tasks, including object removal, object addition, and localized alteration.
UltraEdit~\cite{NEURIPS2024_05a30a0f} proposes a massive $4$M-sample dataset for instruction-based image editing, enabling support for region-based instruction. More recently, Pico-Banana-400k~\cite{picobanana} introduced a $400$K synthetic dataset covering both single- and multi-turn editing, while ImgEdit~\cite{ye2025imgedit} proposed a large-scale counterpart spanning diverse edit types such as background change, motion change, and style transfer. HumanEdit~\cite{human_edit} contributes $5\text{,}751$ manually constructed editing pairs, emphasizing annotation quality through dedicated human annotators.

\noindent\textbf{Positioning of our work.}
Contrary to the works discussed above, we present a benchmark specifically designed to evaluate visual object consistency.
While existing datasets focus on general-purpose editing tasks (\eg, object removal, style transfer, or inpainting), ABO-Edit provides pixel-accurate target images rendered from artist-designed 3D assets, guaranteeing that geometry, texture, and material are preserved by construction.
ABO-Edit requires models to perform a sequence of challenging sub-tasks: disambiguating and selecting a target product from natural language among visually similar candidates within a single image, removing backgrounds, generating physically plausible shadows, and executing 3D rotations with degree-level angular precision.
Furthermore, ABO-Edit additionally offers degree-level orientation annotations (with multi-view samples), enabling systematic evaluation of spatial consistency under 3D transformations, a capability absent from all prior benchmarks.

\section{Do Generative Models Have Spatial Awareness?}
\label{sup:sec:spatial_awareness}
In this section, we provide additional analysis on the spatial awareness of generative models.
\subsection{Fine-grained Rotation Control.}
\label{sup:spatial_Exp:fine_grained_control}
\begin{figure*}[h]
  \centering
  \includegraphics[width=1\linewidth]{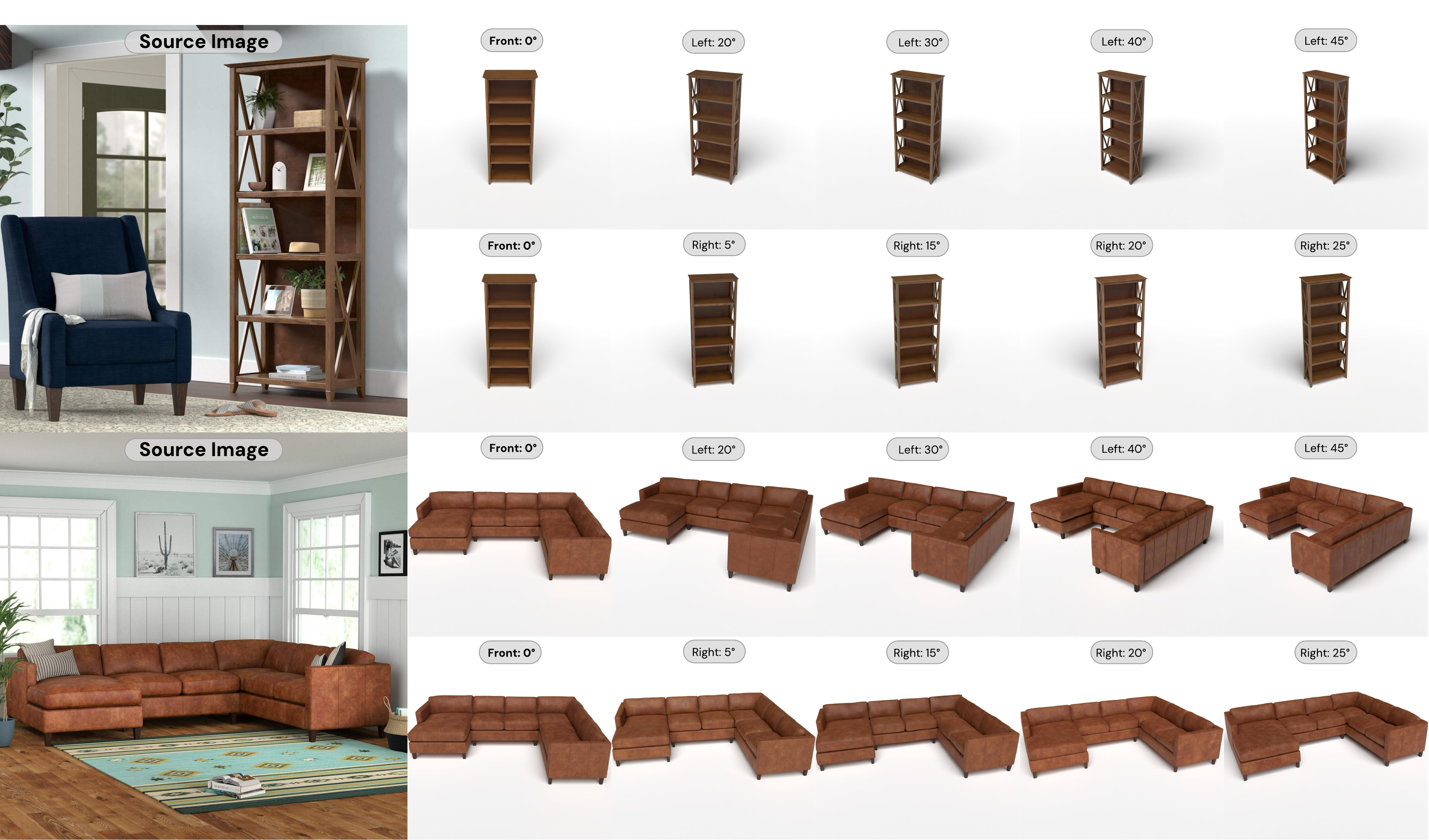}
    \caption{\textbf{Fine-grained rotation control.} Given a source image (leftmost column), our model generates novel views at specified rotation angles while preserving object identity and appearance. Results are shown for two distinct objects: a shelf (top) and a sofa (bottom). For visualization clarity, we display outputs at \SI{5}{\degree} intervals; however, our model supports continuous viewpoint manipulation at \SI{1}{\degree} granularity, enabling precise and consistent control across diverse object categories.}
  \label{fig:sup:rotation_control}
\end{figure*}

Fig.~\ref{fig:sup:rotation_control} illustrates the fine-grained rotation control achieved by our fine-tuned model. The top row shows a brown cabinet as the source image; we then generate novel views at varying angles using our model (based on \texttt{Qwen-Image-Edit-2509}~\cite{qwen_image_edit}) while keeping all other attributes fixed.
The bottom row repeats the experiment with a different source image, \ie a brown sofa.
The results demonstrate precise viewpoint control and strong appearance consistency across generated views, with challenging structure preservation.

\subsection{Azimuth Distribution.}
\label{ssup:spatial_Exp:azimuth}
\begin{figure*}[t]
  \centering
  \begin{subfigure}{0.48\linewidth}
    \centering
    \includegraphics[width=\linewidth, trim=0 0 0 0.5cm, clip]{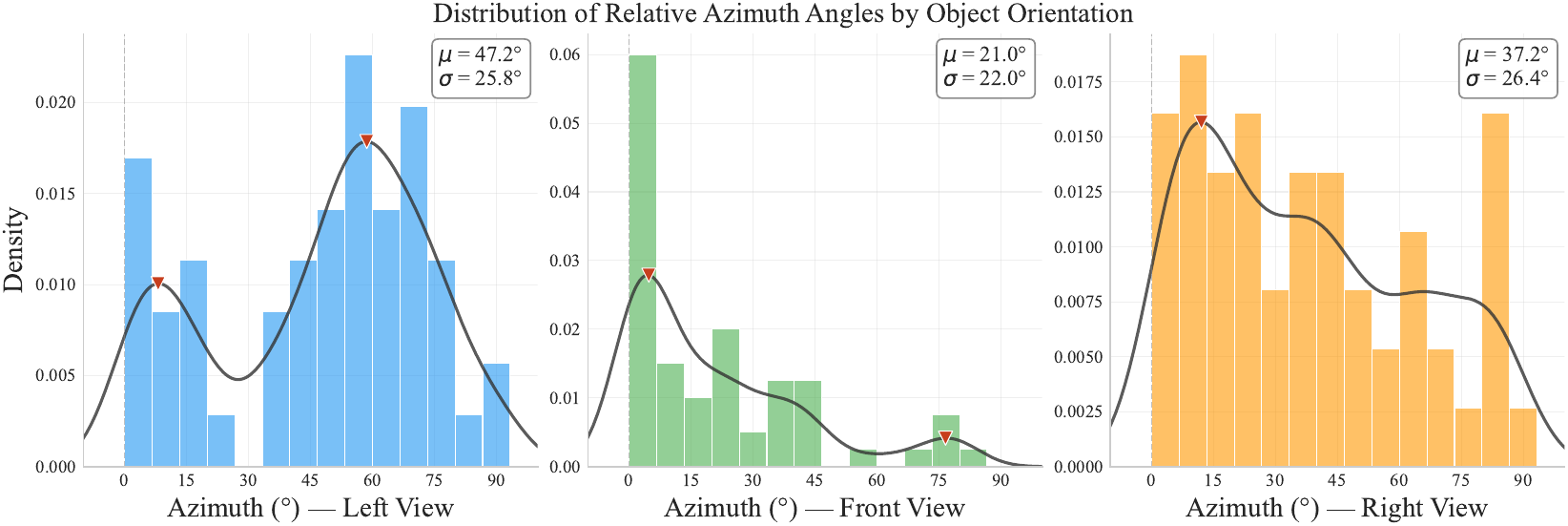}
    \caption{\texttt{Kandinsky}~\cite{kanidsky5} (zero-shot)}
    \label{fig:sup:exp:rotation_distributions_by_bucket_kandisky}
  \end{subfigure}
  \hfill
  \begin{subfigure}{0.48\linewidth}
    \centering
    \includegraphics[width=\linewidth, trim=0 0 0 0.5cm, clip]{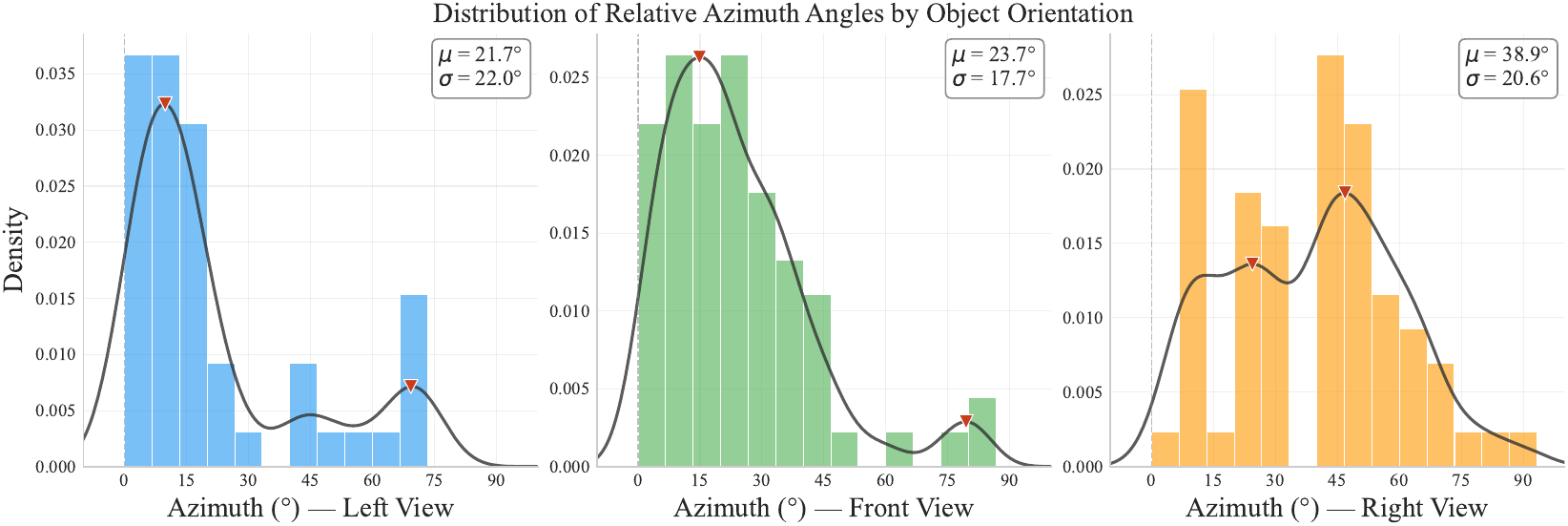}
    \caption{\texttt{FLUX.1}~\cite{flux_1} (zero-shot)}
    \label{fig:sup:exp:rotation_distributions_by_bucket_flux1}
  \end{subfigure}
  
  \vspace{0.5em}
  
  \begin{subfigure}{0.48\linewidth}
    \centering
    \includegraphics[width=\linewidth, trim=0 0 0 0.5cm, clip]{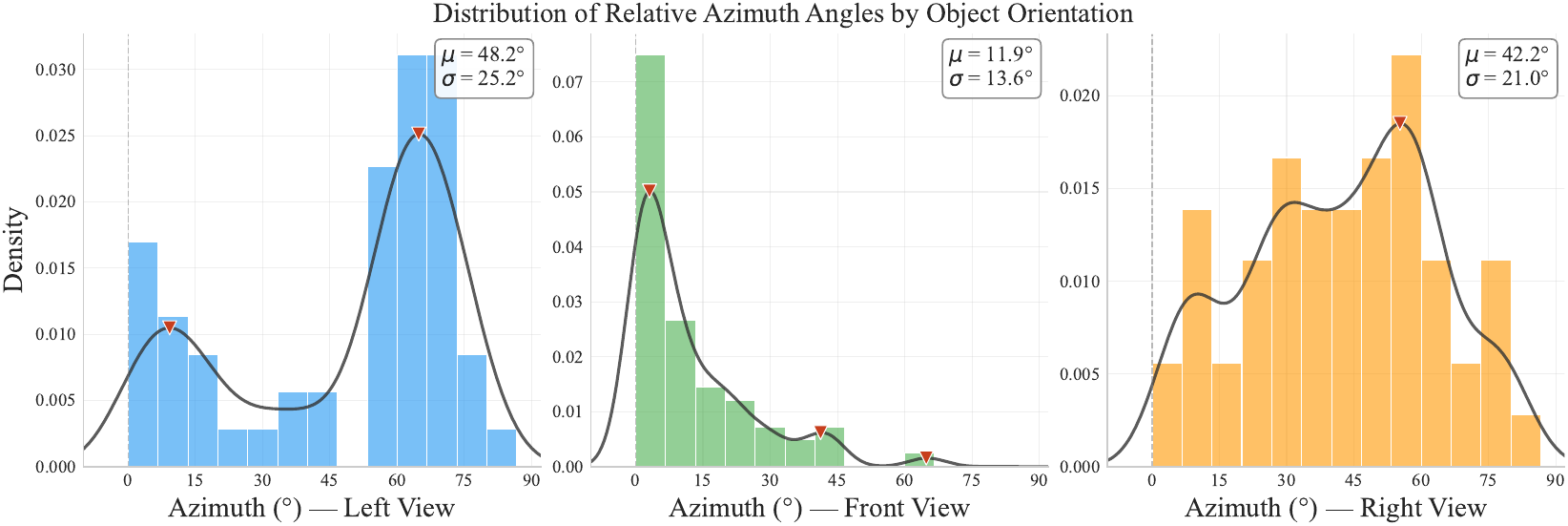}
    \caption{\texttt{FLUX.2}~\cite{flux2} (zero-shot)}
    \label{fig:sup:exp:rotation_distributions_by_bucket_flux2}
  \end{subfigure}
  \hfill
  \begin{subfigure}{0.48\linewidth}
    \centering
    \includegraphics[width=\linewidth, trim=0 0 0 0.5cm, clip]{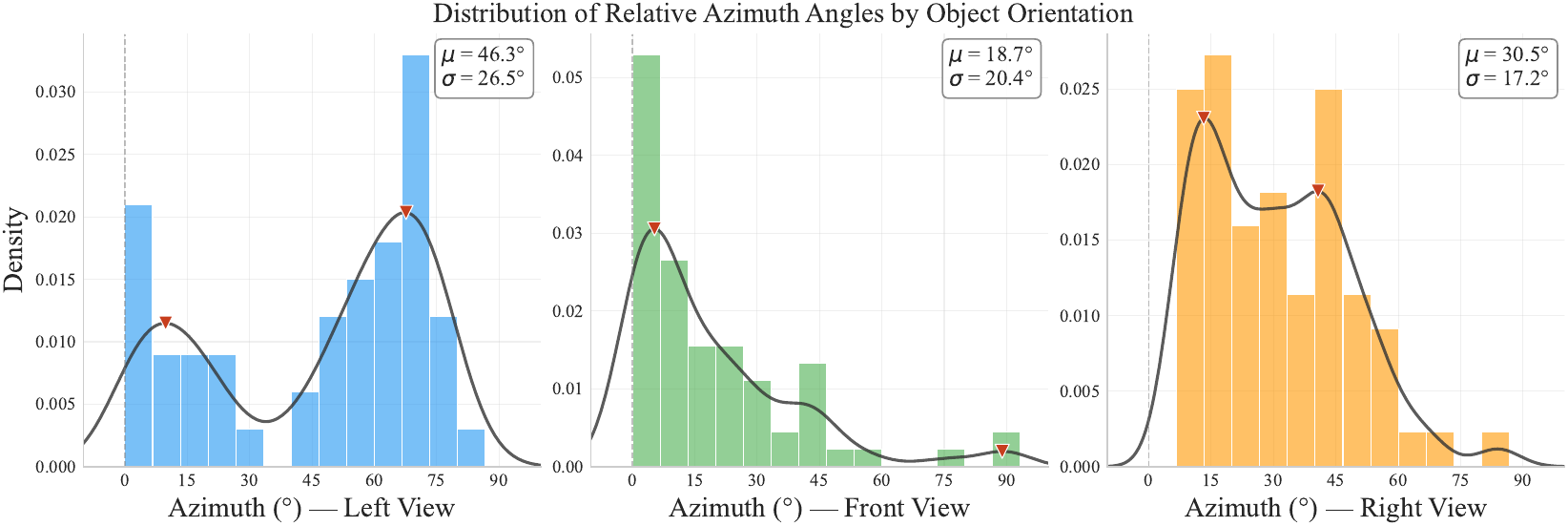}
    \caption{\texttt{Qwen-IE-2511}~\cite{qwen_image_edit} (zero-shot)}
    \label{fig:sup:exp:rotation_distributions_by_bucket_qwen_2511}
  \end{subfigure}
  
  \vspace{0.5em}
  
  \begin{subfigure}{0.48\linewidth}
    \centering
    \includegraphics[width=\linewidth, trim=0 0 0 0.5cm, clip]{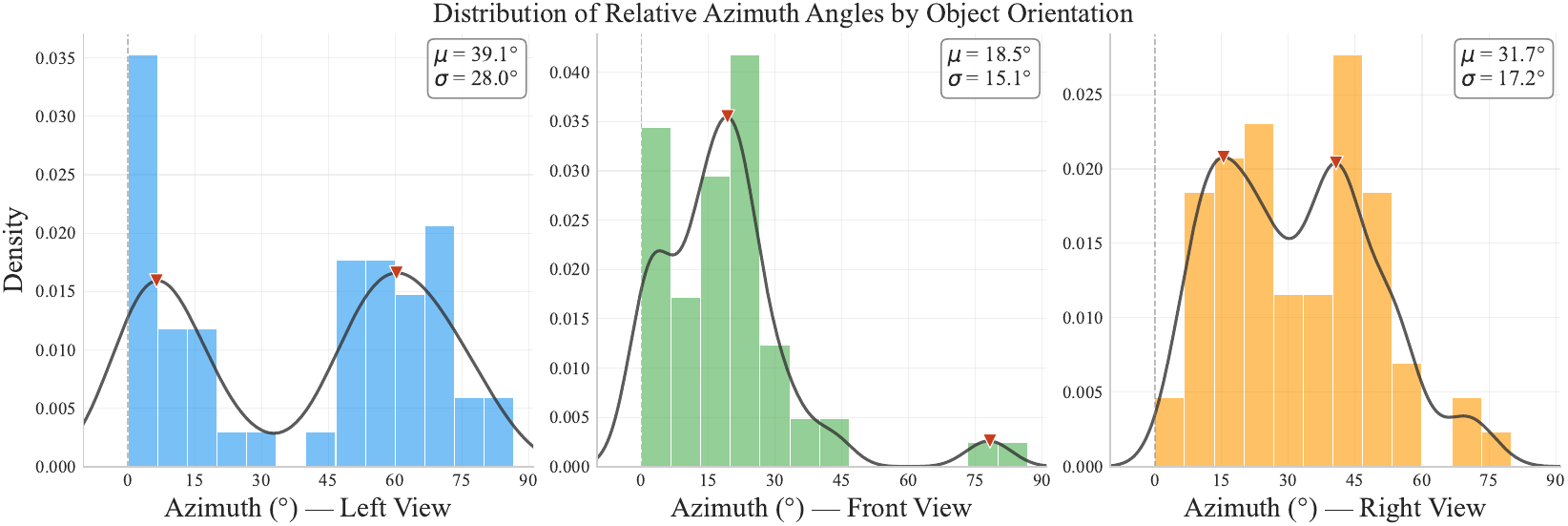}
    \caption{\texttt{Qwen-IE-2509}~\cite{qwen_image_edit} (zero-shot)}
    \label{fig:sup:exp:rotation_distributions_by_bucket_vanilla_qwen_2509}
  \end{subfigure}
  \hfill
  \begin{subfigure}{0.48\linewidth}
    \centering
    \includegraphics[width=\linewidth, trim=0 0 0 0.5cm, clip]{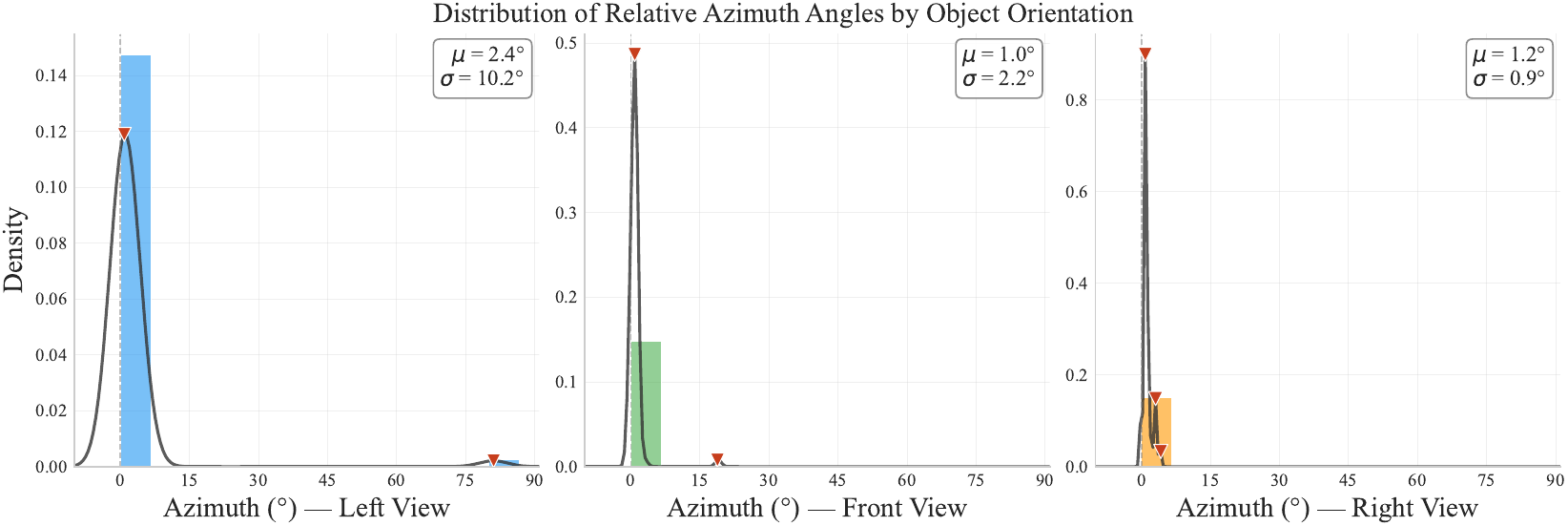}
    \caption{\texttt{Qwen-IE-2509}~\cite{qwen_image_edit} \textbf{(fine-tuned)}}
    \label{fig:sup:exp:rotation_distributions_by_bucket_qwen_finetuned}
  \end{subfigure}

  \caption{Azimuth error distributions across viewpoints (\textit{left}, \textit{front}, \textit{right}) for various models. (a–e)~Zero-shot baselines exhibit broad, often bimodal distributions indicating frequent orientation errors. (f)~Fine-tuning on \datasetName{} yields sharply peaked distributions near zero, demonstrating precise viewpoint control.}
  \label{fig:sup:rotation_distributions_by_bucket}
\end{figure*}

In Fig.~\ref{fig:sup:rotation_distributions_by_bucket}, we report the distribution of azimuth error between the generated and target images, across fine-tuned and zero-shot models.
From the figure, we can derive the following observations:
\begin{enumerate}[label=\textit{(\roman*)}]
    \item Zero-shot models show higher errors on this task: the \textit{left} view yields a mean error of \SI[round-mode=none]{40.5}{\degree}, \textit{front} \SI[round-mode=none]{18.76}{\degree}, and \textit{right} \SI[round-mode=none]{36.1}{\degree}, indicating that fine-grained spatial placement from text prompts alone remains challenging without task-specific adaptation.
    \item For one model~\cite{flux_1}, the \textit{left} view azimuth error is nearly half that of the \textit{right} view, suggesting a potential bias in its training data distribution.
    \item Nearly all models exhibit bimodal error distributions, characterized by high mean azimuth errors in object placement. Furthermore, these models frequently misinterpret the target viewpoint altogether, further underscoring the difficulty of this task without task-specific training.
\end{enumerate}
We ground such errors in visual examples in the following Section.

\subsection{Rotation Failures.}
\begin{figure*}[h]
  \centering
  \includegraphics[width=0.7\linewidth]{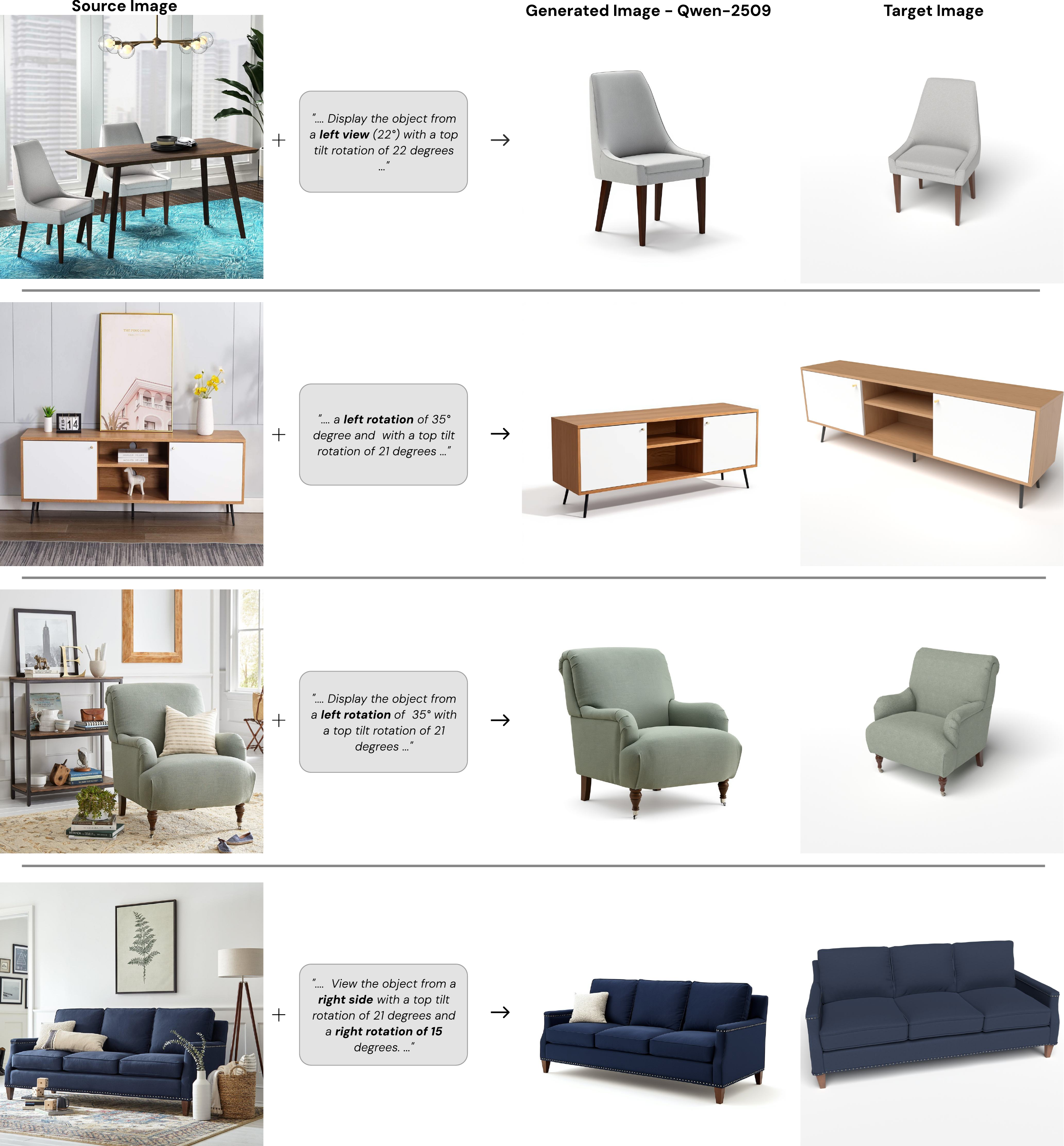}
    \caption{Representative failure cases for spatial reasoning on our validation set. Each triplet shows: source lifestyle image, zero-shot model output~\cite{qwen_image_edit}, and expected target with correct orientation.}
  \label{fig:sup:rotation_failures}
\end{figure*}

Fig.~\ref{fig:sup:rotation_failures} illustrates representative orientation failures. 
When evaluated zero-shot on our validation set, one representative model~\cite{qwen_image_edit} shows limited spatial controllability on this task: left rotation prompts frequently yield right-facing views, and vice versa, suggesting the model lacks a coherent internal representation of 3D orientation.
\label{ssup:spatial_Exp:failures}

\section{Human Evaluation}
\label{supmat:human_eval}
We conducted a blind pairwise comparison between our method and the baseline on $278$ images. Human annotators were asked to select the preferred result, and the final preference was determined by majority vote. Our method was preferred 4.4\% more often than the baseline.
\section{Flow Matching}
\label{supmat:flow_model}
\subsection{$\hatxzero$ Derivation.}

During inference, at any time step $t$, we can retrieve the clean image $\hatxzero$ from the current state $\xt$ via:
$\hatxzero = \xt - t\vPred$.

\noindent\textit{Derivation.}
\label{supmat:x0_derivation}
During training, the ground truth velocity is defined as:
\[
\mathbf{v} = \xone - \xzero.
\]
We start by reporting and then rewriting Eq.~\textbf{\textcolor{red}{2}}:
\begin{align*}
\xt &= (1-t) \xzero + t \xone \\
    &= \xzero + t(\xone - \xzero) \\
    &= \xzero + t\mathbf{v}
\end{align*}
Rearranging for $\xzero$:
\[
\xzero = \xt - t\mathbf{v}
\]
During inference, we do not have access to the ground truth velocity $\mathbf{v}$. 
Instead, we substitute the model's prediction $\vPred \approx \mathbf{v}$, yielding 
the estimated clean image:
\[
\hatxzerot = \xt - t\vPred
\]
Note that this result is consistent with the $n$ steps integration of the ODE.

\end{document}